\pdfoutput=1

\documentclass[11pt]{article}

\usepackage[final]{acl}

\usepackage{times}
\usepackage{latexsym}
\usepackage{times}
\usepackage{enumitem}
\usepackage{amsmath}
\usepackage{amsfonts}
\usepackage{latexsym}
\usepackage{graphicx}
\usepackage{makecell}
\usepackage{multirow}
\usepackage{booktabs}
\usepackage{algorithmic}
\usepackage{algorithm}

\usepackage[T1]{fontenc}

\usepackage[utf8]{inputenc}

\usepackage{microtype}

\title{Golden Touchstone: A Comprehensive Bilingual Benchmark for Evaluating Financial Large Language Models}

\author{
\bf Xiaojun Wu$^{\ast\ 1,2}$,
Junxi Liu$^{\ast\ 1,5}$,
Huanyi Su$^{1,4}$,
Zhouchi Lin$^{1}$,
Yiyan Qi$^{1}$, 
Chengjin Xu$^{1,6}$, \\
\bf Jiajun Su$^{1}$,
Jiajie Zhong$^{1}$,
Fuwei Wang$^{1}$,
Saizhuo Wang$^{1,3}$, 
Fengrui Hua$^{1,2}$, \\
\bf Jia Li$^{\dagger\ 2}$,
Jian Guo$^{\dagger\ 1}$
\\
\\
$^{1}$IDEA Research \quad
$^{2}$The Hong Kong University of Science and Technology (Guangzhou) \\
$^{3}$The Hong Kong University of Science and Technology \\
$^{4}$Nanjing University \quad
$^{5}$South China Normal University \quad
$^{6}$DataArcTech Ltd.
}

\begin{document}
\maketitle
{
  \renewcommand{\thefootnote}%
    {\fnsymbol{footnote}}
  \footnotetext[1]{Equal Contribution: wuxiaojun@idea.edu.cn, liujunxi@m.scnu.edu.cn}
  \footnotetext[2]{Corresponding Author: jialee@connect.hkust-gz.edu.cn, guojian@idea.edu.cn}
}
\begin{abstract}
As large language models (LLMs) increasingly permeate the financial sector, there is a pressing need for a standardized method to comprehensively assess their performance. Existing financial benchmarks often suffer from limited language and task coverage, low-quality datasets, and inadequate adaptability for LLM evaluation. To address these limitations, we introduce Golden Touchstone, a comprehensive bilingual benchmark for financial LLMs, encompassing eight core financial NLP tasks in both Chinese and English. Developed from extensive open-source data collection and industry-specific demands, this benchmark thoroughly assesses models' language understanding and generation capabilities. Through comparative analysis of major models such as GPT-4o, Llama3, FinGPT, and FinMA, we reveal their strengths and limitations in processing complex financial information. Additionally, we open-source Touchstone-GPT, a financial LLM trained through continual pre-training and instruction tuning, which demonstrates strong performance on the bilingual benchmark but still has limitations in specific tasks. This research provides a practical evaluation tool for financial LLMs and guides future development and optimization.
The source code for Golden Touchstone and model weight of Touchstone-GPT have been made publicly available at \url{https://github.com/IDEA-FinAI/Golden-Touchstone}.
\end{abstract}

\label{sec:introduction}
\section{Introduction}
\begin{figure*}[htbp]
    \centering
    \includegraphics[width=0.95\textwidth]{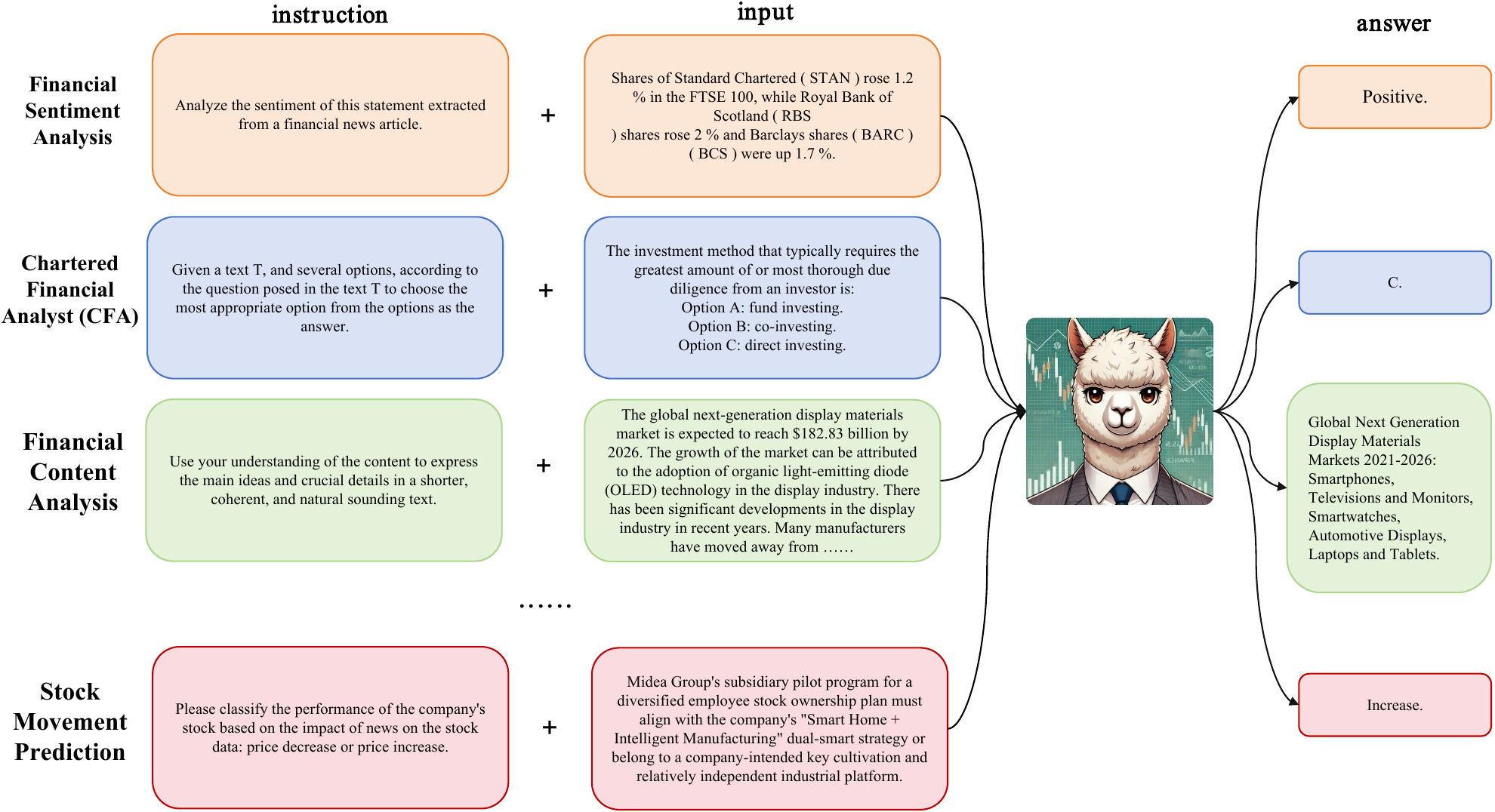}
    \caption{Financial large language models are designed to perform specialized tasks such as financial sentiment analysis, content analysis, stock movement prediction, and financial analyst level question answering by interpreting and processing structured instructions and various input data to generate precise outputs.}
    \label{fig:Overview_FinLLM}
\end{figure*}

The rapid development of both proprietary \citep{brown2020gpt3, ouyang2022instruct_gpt, openai2023gpt4, anthropic2024claude, team2023gemini} and open-source Large Language Models (LLMs) \citep{touvron2023llama, touvron2023llama2, llama3, bai2023qwen, yang2024qwen2, deepseekv2, young2024yi, zeng2023glm-130b, baichuan2023baichuan2, gan2023ziya2, zhang2022fengshenbang} has led to their increasing application in various fields, including finance \citep{wu2023bloomberggpt, lopez2023chatgpt_stock_move}, healthcare \citep{thirunavukarasu2023llm_med, tian2023chimed}, and law \citep{cui2023chatlaw, xiao2021lawformer}. Among these, the financial sector shown in Figure.\ref{fig:Overview_FinLLM} stands out as a critical area for LLM application due to its rich textual information and high practical value.

In recent years, a variety of advanced financial large language models (FinLLMs) have emerged, capable of specialized tasks such as financial sentiment analysis, content summarization, stock movement prediction, and question answering \citep{yang2023fingpt, xie2023pixiu, li2023cfgpt, chen2023disc, zhang2023xuanyuan}. These models leverage unique frameworks and tuning methods to enhance their performance on domain-specific benchmarks, offering robust solutions for real-world financial applications. However, existing financial benchmarks often suffer from limited language and task coverage, low-quality datasets, and inadequate adaptability for LLM evaluation, leading to poor evaluation results \citep{shah2022flue, lu2023bbt, xie2023pixiu, xie2024finben, yang2023fingpt, lei2023cfbenchmark, zhang2023fineval}.

To address these challenges, we propose Golden Touchstone, a comprehensive bilingual benchmark for financial LLMs, encompassing eight core financial NLP tasks in both Chinese and English. Golden Touchstone provides high-quality datasets, task-aligned metrics, and instructional templates to guide LLMs in generating task-appropriate responses. We evaluated several state-of-the-art models, including GPT-4o, Qwen-3, Llama-3, FinGPT, and FinMA, on this benchmark. Results indicate that while these models perform well on tasks such as sentiment analysis and entity extraction, there is significant room for improvement in areas like stock movement prediction and classification tasks. Additionally, we open-source Touchstone-GPT, a financial LLM trained through domain-specific continual pre-training and instruction tuning, which serves as a new baseline for future research.

Our contributions are as follows:
\begin{itemize}
\item Introduction of Golden Touchstone, a comprehensive bilingual benchmark for financial LLMs, encompassing 22 datasets across eight tasks in both Chinese and English.
\item Evaluation of state-of-the-art LLMs and FinLLMs on Golden Touchstone, highlighting their strengths and limitations across various tasks.
\item Open-sourcing of Touchstone-GPT, a financial LLM trained through domain-specific continual pre-training and instruction tuning, fostering further advancements in financial AI.
\end{itemize}

\section{Benchmark Design}

\subsection{Current Benchmark Status}
Existing open-source financial benchmarks have made significant strides in evaluating financial natural language processing (NLP) tasks. FLUE \citep{shah2022flue} pioneered English financial NLP evaluation, covering sentiment analysis, news classification, and other critical tasks. Subsequently, PIXIU \citep{xie2023pixiu} and FinBen \citep{xie2024finben} expanded task coverage, while in the Chinese domain, BBT-Benchmark \citep{lu2023bbt} introduced the first comprehensive Chinese financial evaluation framework. 

These benchmarks suffer from several critical limitations. Firstly, there is inconsistent data quality across different tasks, as shown in Table \ref{tab:analysis_tasks}. Additionally, large language models face challenges in numerical understanding, a problem highlighted in various studies \citep{shen2023positional, akhtar2023numerical_reasoning, schwartz2024numerologic, hu2024no}. Lastly, the benchmarks lack bilingual assessment capabilities, a gap that has been pointed out in recent research \citep{nie2024cfinbench, lei2023cfbenchmark}. These limitations collectively hinder the overall effectiveness and reliability of the benchmarks.
To address these limitations, we propose a unified benchmark that integrates high-quality financial datasets from both English and Chinese domains. Our approach aims to provide a more comprehensive and linguistically diverse evaluation of financial large language models (LLMs).

\begin{table*}[htbp]
\centering
\caption{Diversity of Financial Analysis Tasks Across Different Benchmarks}
\label{tab:analysis_tasks}
\fontsize{7pt}{7pt}\selectfont
\begin{tabular}{@{}lcccccccc@{}}
\toprule
Benchmarks & Sent. Anal. & Classif. & Ent. Extr. & Rel. Extr. & Multi. Choice & Summ. & Quest. Ans. & Stock Pred. \\
\midrule
FinGPT-Bench \citep{wang2023fingptbench} & $\checkmark$ & $\checkmark$ & $\checkmark$ & $\checkmark$ &  &  &  &  \\
FinBen \citep{xie2024finben} & $\checkmark$ & $\checkmark$ & $\checkmark$ & $\checkmark$ &  & $\checkmark$ & $\checkmark$ & $\checkmark$ \\
BBT-Fin \citep{lu2023bbt} & $\checkmark$ & $\checkmark$ & $\checkmark$ & $\checkmark$ &  & $\checkmark$ & $\checkmark$ &  \\
Fin-Eval \citep{zhang2023fineval} &  &  &  &  & $\checkmark$ &  &  &  \\
CFBenchmark \citep{lei2023cfbenchmark} & $\checkmark$ & $\checkmark$ & $\checkmark$ &  &  & $\checkmark$ & $\checkmark$ &  \\
\midrule
Golden-Touchstone & $\checkmark$ & $\checkmark$ & $\checkmark$ & $\checkmark$ & $\checkmark$ & $\checkmark$ & $\checkmark$ & $\checkmark$ \\
\bottomrule
\end{tabular}
\end{table*}

\subsection{Golden Touchstone Benchmark Framework}
\begin{figure*}[htbp]
    \centering
    \includegraphics[width=1\textwidth]{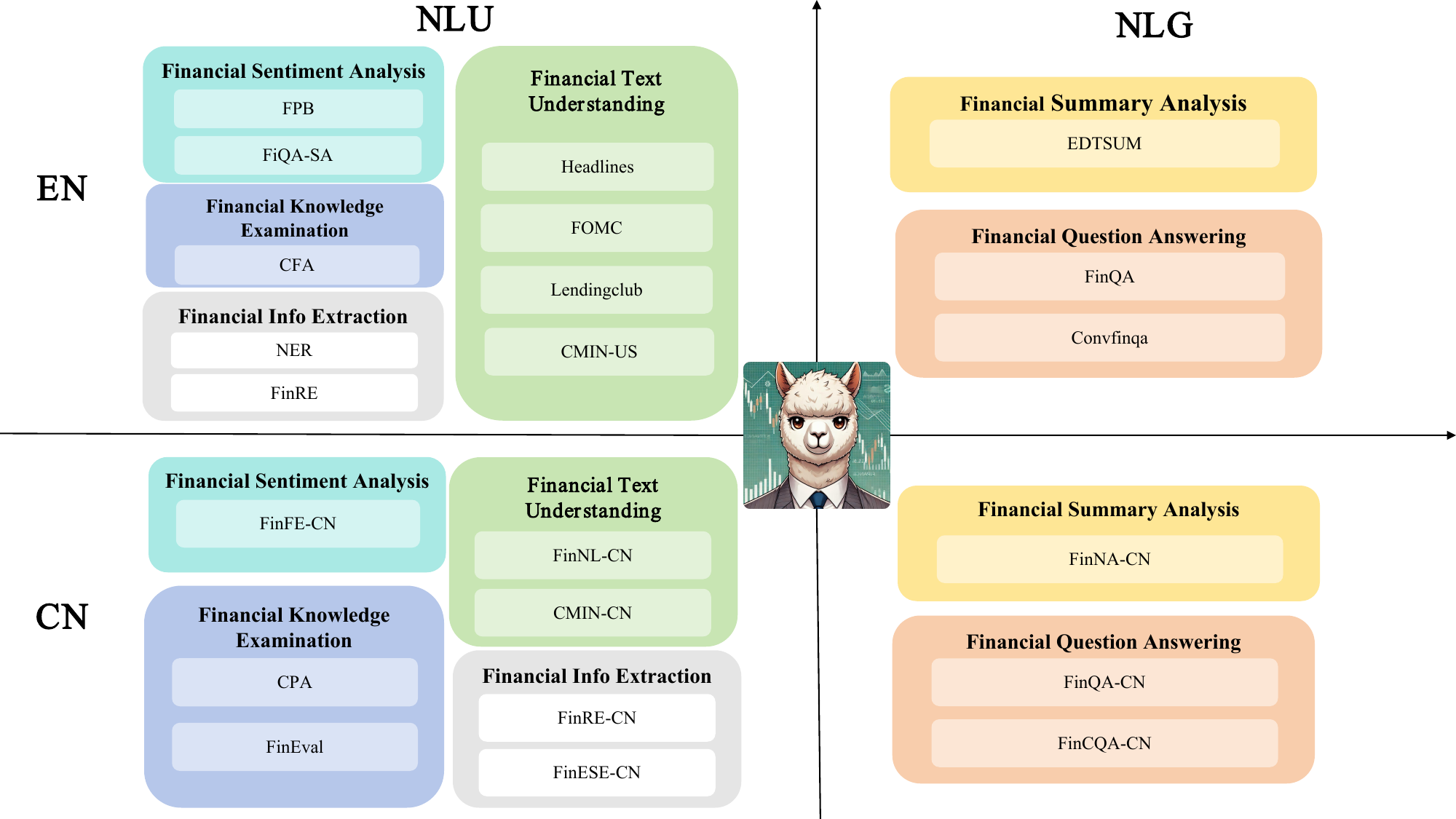}
    \centering \caption{Financial NLP tasks are categorized along two dimensions: task types, divided into financial NLU (Natural Language Understanding) and financial NLG (Natural Language Generation), and language, categorized as English and Chinese. We organized the collected high-quality datasets along these axes. }
    \label{fig:Benchmark_Touchstone}
\end{figure*}

Addressing these critical limitations, we introduce the Golden Touchstone benchmark, a comprehensive bilingual evaluation framework designed to holistically assess financial language models. Conceptualized around two primary dimensions—task types and language coverage—our approach represents a significant departure from existing evaluation methodologies. The overview framework can be seen in Figure.\ref{fig:Benchmark_Touchstone}.
Firstly, our Golden Touchstone benchmark categorizes financial NLP tasks across two dimensions: the first dimension is task type including Natural Language Understanding (NLU) including sentiment analysis, knowledge examination, information extraction, text understanding tasks and Natural Language Generation (NLG) including summary analysis, question answering tasks, and the second dimension is language including English and Chinese.

The benchmark strategically integrates English and Chinese datasets across eight sophisticated sub-tasks, spanning Natural Language Understanding (NLU) and Natural Language Generation (NLG). By carefully curating high-quality datasets from existing benchmarks and datasets. The detailed data statistics are shown in Appendix.\ref{sec:data_statistics}. The benchmark encompasses eight critical sub-tasks:
\begin{itemize}
   \item \textbf{Sentiment Analysis}: Utilizing datasets like FPB \citep{malo2014fpb} and FiQA-SA \citep{maia2018finqa_sa} and FinFE-CN \citep{lu2023bbt}
   \item \textbf{Classification}: Integrating Headlines \citep{sinha2021headlines}, FOMC \citep{shah2023fomc}, and LendingClub \citep{feng2023lendingclub} and FinNL-CN \citep{lu2023bbt} datasets
   \item \textbf{Entity Recognition}: Using NER \citep{alvarado2015NER} and FinESE-CN \citep{lu2023bbt} datasets
   \item \textbf{Relation Extraction}: Incorporating FinRED \citep{sharma2022finred} and FinRE-CN \citep{lu2023bbt} datasets
   \item \textbf{Multiple Choice}: Drawing from CFA \citep{yang2024ideafinbench} and FinEval \citep{zhang2023fineval} and CPA \citep{yang2024ideafinbench} datasets
   \item \textbf{Summarization}: Employing EDTSUM \citep{zhou2021edtsum} and FinNA-CN \cite{lu2023bbt} datasets
   \item \textbf{Question Answering}: Utilizing FinQA \citep{chen2021finqa} and FinQA-CN and FinCQA-CN \citep{lu2023bbt} datasets
   \item \textbf{Stock Movement Prediction}: Introducing news-based prediction using CMIN-US and CMIN-CN \citep{luo2023cmin} datasets
\end{itemize}

A implementation consideration involves the treatment of [unknown] labels across various tasks. Our approach consistently treats [unknown] labels as incorrect predictions, which maintains uniform evaluation standards across all tasks, avoids potential biases introduced by alternative methods, and reflects real-world scenarios where models must make definitive predictions. This approach proves superior to alternative methods such as sample omission, random assignment, or fixed selection, which may introduce evaluation biases or distort performance metrics.
The benchmark also design incorporates several features to ensure meaningful cross-lingual comparisons, including maintaining identical task structures across English and Chinese benchmarks, using comparable evaluation metrics for equivalent tasks, and choosing datasets with similar characteristics and difficulty levels.
These design choices enable robust comparison of model performance across languages while accounting for linguistic and cultural differences in financial contexts.

By addressing previous benchmarks' limitations, our approach provides a more robust, comprehensive, and linguistically diverse framework for evaluating financial large language models. The benchmark not only expands task coverage but also addresses critical challenges in current financial NLP evaluation methodologies.
Our key methodological innovations reflect significant efforts in data selection, cleaning, and organization. We replaced time-series tabular data with news-based stock prediction to enhance the relevance and diversity of the dataset. Additionally, we ensured bilingual task alignment to address the lack of multilingual assessment capabilities, while carefully selecting datasets with consistent and high-quality labeling to overcome issues related to data quality. Furthermore, we developed a unified evaluation approach across different financial NLP tasks to ensure robust and comparable results. In line with our commitment to fostering research transparency and collaboration, we have not only open-sourced the benchmark datasets but also released a well-documented, clear, reproducible, and extensible evaluation code framework for the broader research community to utilize and build upon. 
The Golden Touchstone benchmark represents a significant step forward in financial NLP evaluation, addressing key limitations of existing benchmarks while providing a comprehensive, bilingual framework for assessing model performance. Its systematic design, rigorous evaluation methodology, and practical focus make it an essential tool for researchers and practitioners in financial NLP.


\section{Experiments}

\subsection{Experimental Setup}

\textbf{Baselines.}
We conducted an extensive experimental evaluation against the Golden Touchstone Benchmark, incorporating a comprehensive array of models. For all models and inference tasks, we set the PyTorch and CUDA random seeds and configured the model with a greedy decoding strategy. This ensures reproducibility of experimental results and eliminates the influence of sampling decoding strategies on the final generated outputs. This included cutting-edge commercial models such as GPT-4o \citep{openai2023gpt4}, alongside prominent open-source alternatives like Meta Llama-3 \citep{llama3} and Alibaba Qwen-3 \citep{yang2025qwen3}. Additionally, we integrated the latest and most influential financial language models (FInLLMs), namely FinGPT \citep{yang2023fingpt}, FinMA \citep{xie2024finben}, CFGPT \citep{li2023cfgpt}, and DISC-FinLLM \cite{chen2023disc}. These models were meticulously selected to represent a diverse spectrum of capabilities, ranging from general-purpose language understanding to specialized financial domain expertise. Our experiments aimed to rigorously assess the performance, robustness, and adaptability of each model within the context of financial data processing and analysis. The results provide valuable insights into the strengths and limitations of current state-of-the-art models, offering a foundation for future advancements in financial language modeling.

\textbf{Touchstone-GPT Training.}
To further contribute to the research and development of FInLLMs and Financial benchmarks for LLMs, we have meticulously trained and open-sourced a Touchstone-GPT model. This initiative aims to serve as a valuable resource for advancing the field, providing a robust and versatile model that can be utilized for a wide range of financial language tasks. 
We adopted a two-stage training strategy comprising continuous pre-training and post-training, based on the Qwen-2.5 \citep{yang2024qwen2} foundational model. During the continuous pre-training phase, we initially conducted pre-training on a high-quality  financial corpus containing 100 billions tokens, which included textbooks, encyclopedias, research reports, news articles, and real-time analysis, all meticulously cleaned. In the post-training phase, we employed a standard instruction fine-tuning strategy, collecting, cleaning, and formatting a high-quality dataset of 300,000 instruction-response pairs shown in Tabel.\ref{tab:benchmark_en} and Table.\ref{tab:benchmark_cn}. To avoid catastrophic forgetting in general tasks, we also incorporated general-domain pre-training corpora \citep{gan2023ziya2} and instruction-tuning corpora \citep{peng2023Alpaca_gpt4} into continuous pre-training and post-training. This culminated in the final Touchstone-GPT model.
We utilized Megatron\citep{shoeybi2019megatron} for continuous pre-training and LlamaFactory\citep{zheng2024llamafactory} for instruction post-training as our training frameworks, respectively.
In this study, we employ an advanced model training setup using the AdamW optimizer \citep{kingma2014adam} with a learning rate of 1.0e-5, cosine annealing scheduler \citep{szegedy2016cosine}, and a 10\% warmup ratio \citep{he2016resnet,goyal2017warmup} to enhance training stability and convergence. We enable gradient accumulation \citep{shoeybi2019megatron} and checkpointing \citep{chen2016gradient_ckpt} to simulate larger batch sizes and reduce memory footprint. Training is conducted in mixed bfloat16  precision \citep{micikevicius2017mixed,wang2019bf16} with DeepSpeed's ZeRO-1 optimization \citep{rajbhandari2020zero}, reducing memory consumption and allowing for larger model training. 
This comprehensive setup optimizes efficiency and performance, providing an effective solution for large-scale deep learning model training.

Our training was conducted on 4 NVIDIA DGX servers, each equipped with 8 A100 GPUs, and spanned a period of 4 weeks. Inference was performed on a single NVIDIA DGX server with eight A100 GPUs, utilizing parallel batch inference. During the pre-training phase, we employed a data packing strategy \citep{krell2021packing} and batch dynamic right padding strategy in the instruction tuning phase \citep{wolf-etal-2020-transformers}, while the inference phase incorporated a batch left padding strategy \citep{wolf-etal-2020-transformers}.

\subsection{Evaluation Results}

In this section, we provide a detailed analysis of the evaluation and results for both the English and Chinese benchmarks. We discuss task-specific performances and identify key areas of strengths and weaknesses for each model. The following sections present insights for the English and Chinese benchmarks, each highlighting the differences in model capabilities across a variety of NLP tasks.


In this overall analysis, we synthesize insights from both the English and Chinese benchmarks by evaluating the strengths and weaknesses of each model, as well as examining the performance across different types of tasks and datasets.

\begin{figure*}
    \centering
    \includegraphics[width=\linewidth]{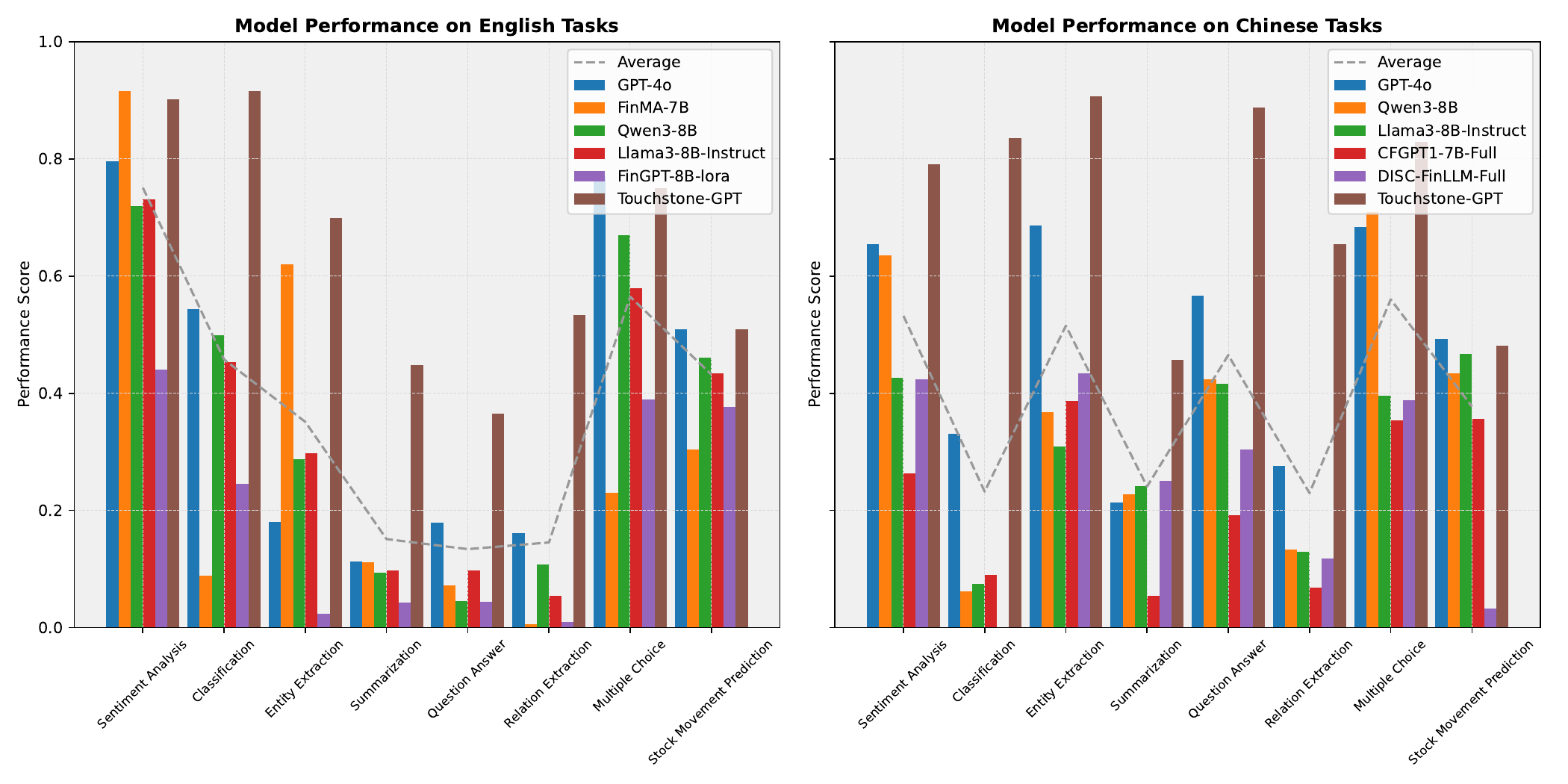}
    \caption{Comparison of performance from the perspective of models. 
Each subplot represents the performance of a models on both English and Chinese tasks. 
The bars indicate the model's performance on each task, while the dashed gray line represents the average performance across all models for that task. }
    \label{fig:model_task_performance_comparison}
\end{figure*}

From the perspective of individual models in Figure.\ref{fig:model_task_performance_comparison}, \textbf{GPT-4o} shows strong performance in sentiment analysis and structured tasks like multiple choice, indicating robust general language understanding capabilities. However, its weakness lies in relation extraction and detailed entity extraction, which require detailed understanding of complex financial relations. \textbf{FinMA-7B} stands out in sentiment tasks but lacks versatility, especially in question answering and summarization, likely due to the absence of targeted training for diverse NLP challenges. \textbf{Qwen-3-8B} demonstrates a balanced yet still improvable performance profile: it achieves decent results in sentiment analysis tasks but shows notable deficiencies in more complex tasks like question answering and summarization, indicating the need for more targeted domain adaptation training to enhance its specialized capabilities. \textbf{Llama-3-8B-Instruct} excels in english NLU tasks, but shows limitations in tasks requiring chinese tasks, such as entity and relation analysis. The metrics of \textbf{FinGPT-8B-lora} indicating that the current level of domain-specific tuning is insufficient for complex financial tasks. Finally, \textbf{DISC-FinLLM-Full} and \textbf{CFGPT1-7B-Full} demonstrate moderate strengths in entity extraction tasks but lack the robustness needed for broader NLP capabilities, revealing significant gaps in financial language comprehension.

\begin{figure*}
    \centering
    \includegraphics[width=\linewidth]{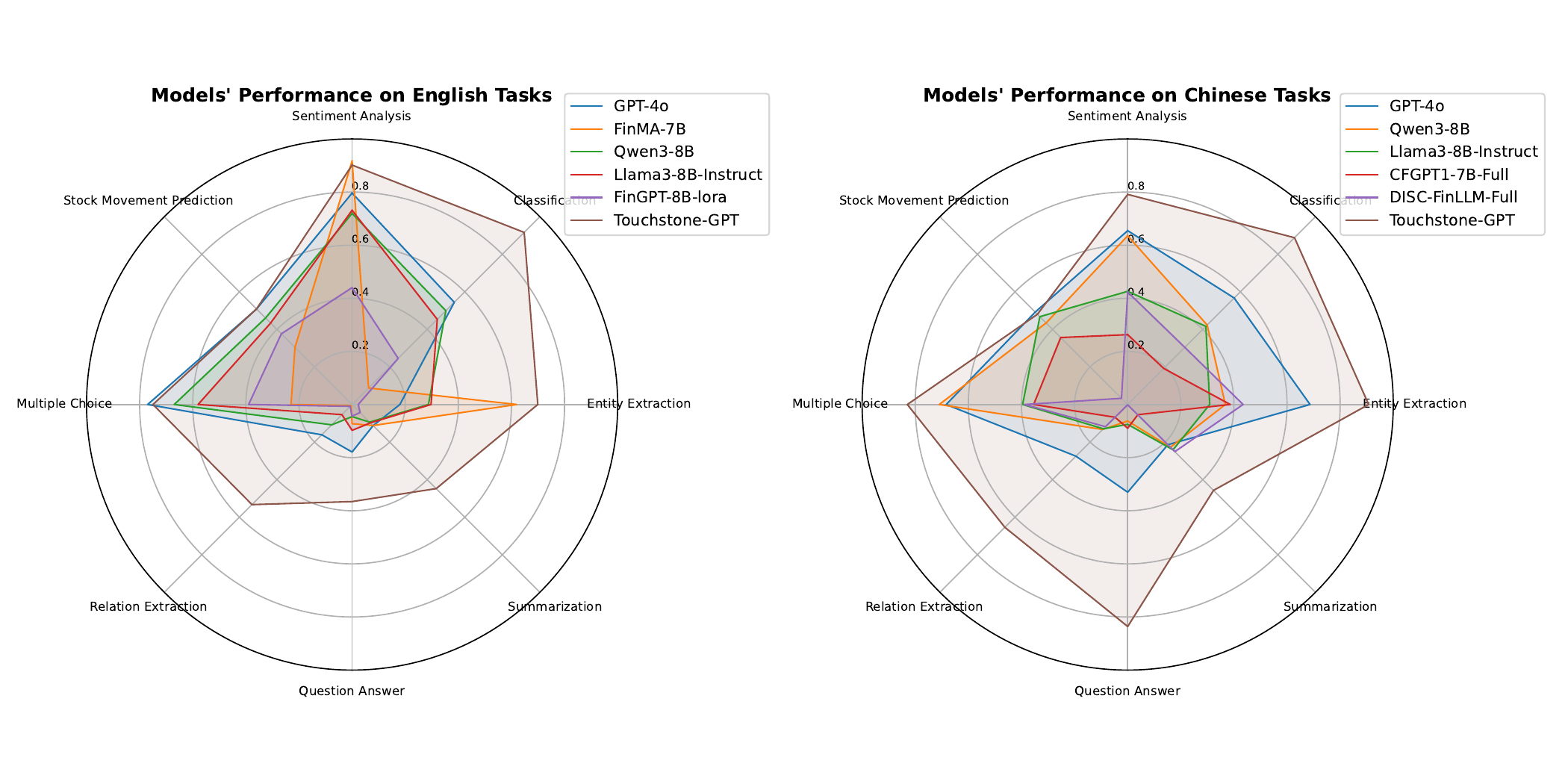}
    \caption{Comparison of performance from the perspective of tasks, illustrating average performance for English and Chinese tasks respectively.}
    \label{fig:radar_chart}
\end{figure*}

From a task perspective in Figure.\ref{fig:radar_chart}, we observe that \textbf{Sentiment Analysis} generally yields high scores across most models, particularly for the English benchmark, indicating that sentiment understanding, even in financial contexts, is relatively well addressed by these models. In contrast, \textbf{Relation Extraction} and \textbf{Question
Answer} in financial domain exhibit notably lower performance, especially for the Chinese benchmark. These results suggest that capturing financial relationships and classifying detailed financial statements pose greater challenges, requiring more sophisticated training datasets or better model architectures. The LendingClub dataset in \textbf{Classification} is a specialized dataset in the field of risk control, requiring more targeted fine-tuning to achieve good results. \textbf{Stock Movement Prediction} also shows low performance across most models, with only a few models such as \textbf{GPT-4o} demonstrating relatively moderate performance, but it is still practically unusable, highlighting the inherent difficulty of this task. Market prediction relying solely on news information is likely insufficient; volume-price data and factor analysis can provide more comprehensive information. However, current large language models are unable to process these inputs, which is a significant area of future research. \textbf{Summarization} also stands out as a weak area for most models, with consistently low BLEU and Rouge scores, reflecting the challenges in generating concise, coherent summaries of complex financial text.

Overall, the insights suggest that while models like \textbf{GPT-4o}, \textbf{FinMA-7B}, and \textbf{Touchstone-GPT} have particular strengths in sentiment analysis and some structured tasks, the overall capability to handle comprehensive financial NLP tasks remains limited. Most models require targeted improvements, especially for relation extraction, summarization, question answering and stock movement prediction in both English and Chinese contexts. This calls for more domain-specific training and the development of specialized datasets that focus on capturing the detailed and often complex financial language, which is crucial for advancing the performance of financial large language models. Furthermore, while \textbf{Touchstone-GPT} demonstrates competitive performance across various tasks due to its robust pre-training and instruction tuning, ongoing refinements and specialized tuning efforts are needed to address specific deficiencies observed in tasks such as summarization, relation extraction, question answering and stock movement prediction.

\begin{table*}
    \centering
    \caption{Performance metrics of financial large language models across English tasks like Sentiment Analysis, Classification, and Summarization. Models include GPT-4o, Llama-3-8B, Qwen-3-8B, FinMA-7B, FinGPT-8B, Mistral-7B, and Touchstone-GPT. The best results of each dataset are marked in \textbf{bold}.}
    \renewcommand\arraystretch{1}
    \scriptsize
    \begin{tabular}{c@{}c@{}c@{}c@{}c@{}c@{}c@{}c@{}c@{}c@{}c@{}c}
        \toprule
        \multirow{1}{*}{\textbf{Task}} & \multicolumn{1}{c}{\textbf{Dataset}} & \multicolumn{1}{c}{\textbf{Metrics}} & \multicolumn{1}{c}{\textbf{\makecell{GPT-4o}}} & \multicolumn{1}{c}{\textbf{\makecell{FinMA-7B\\full}}}  & \multicolumn{1}{c}{\textbf{\makecell{Qwen-3-8B}}} & \multicolumn{1}{c}{\textbf{\makecell{Llama-3-8B\\Instruct}}}  & \multicolumn{1}{c}{\textbf{\makecell{FinGPT-8B\\lora}}} & \multicolumn{1}{c}{\textbf{\makecell{Mistral-7B}}} & \multicolumn{1}{c}{\textbf{\makecell{Touchstone\\GPT}}}\\
         \midrule
         \multirow{4}{*}{\makecell{Sentiment\\Analysis}} & \multirow{2}{*}{FPB} & \multirow{1}{*}{Weighted-F1} & \multirow{1}{*}{0.8084}  & \multirow{1}{*}{\textbf{0.9400}} & \multirow{1}{*}{0.8100} & \multirow{1}{*}{0.7631} & \multirow{1}{*}{0.2727}  & \multirow{1}{*}{0.8110} & \multirow{1}{*}{0.8576}\\

         &&ACC  & \multirow{1}{*}{0.8093} & \multirow{1}{*}{\textbf{0.9402}} & \multirow{1}{*}{0.8020} & \multirow{1}{*}{0.7660} & \multirow{1}{*}{0.3072}   & \multirow{1}{*}{0.8030} & \multirow{1}{*}{0.8557}\\

         &\multirow{2}{*}{Fiqa-SA} & \multirow{1}{*}{Weighted-F1} & \multirow{1}{*}{0.8106}  & \multirow{1}{*}{0.8370} & \multirow{1}{*}{0.6800} & \multirow{1}{*}{0.7515} & \multirow{1}{*}{0.5885} & \multirow{1}{*}{0.6810} & \multirow{1}{*}{\textbf{0.8591}}\\

         &&\multirow{1}{*}{ACC} & \multirow{1}{*}{0.7702}  & \multirow{1}{*}{0.8340}  & \multirow{1}{*}{0.6000} & \multirow{1}{*}{0.7064} & \multirow{1}{*}{0.5872} & \multirow{1}{*}{0.6010} & \multirow{1}{*}{\textbf{0.8638}}\\

         \multirow{6}{*}{\makecell{Classification}} & \multirow{2}{*}{Headlines} & \multirow{1}{*}{Weighted-F1} & \multirow{1}{*}{0.7857}  & \multirow{1}{*}{0.9739} & \multirow{1}{*}{0.7300} & \multirow{1}{*}{0.7006} & \multirow{1}{*}{0.4516}  & \multirow{1}{*}{0.7310} & \multirow{1}{*}{\textbf{0.9866}}\\

         &&\multirow{1}{*}{ACC} & \multirow{1}{*}{0.7931}  & \multirow{1}{*}{0.9739} & \multirow{1}{*}{0.7270} & \multirow{1}{*}{0.7004} & \multirow{1}{*}{0.4331} & \multirow{1}{*}{0.7280} & \multirow{1}{*}{\textbf{0.9866}}\\

         &\multirow{2}{*}{FOMC} & \multirow{1}{*}{Weighted-F1} & \multirow{1}{*}{0.6603} & \multirow{1}{*}{0.3988} & \multirow{1}{*}{0.6150} & \multirow{1}{*}{0.4904} & \multirow{1}{*}{0.2758}  & \multirow{1}{*}{0.6160} & \multirow{1}{*}{\textbf{0.8788}}\\

         &&\multirow{1}{*}{ACC} & \multirow{1}{*}{0.6794} & \multirow{1}{*}{0.4274}  & \multirow{1}{*}{0.6230} & \multirow{1}{*}{0.5625} & \multirow{1}{*}{0.2702} & \multirow{1}{*}{0.6240} & \multirow{1}{*}{\textbf{0.8790}}\\

         & \multirow{2}{*}{lendingclub} & \multirow{1}{*}{Weighted-F1} & \multirow{1}{*}{0.6730} & \multirow{1}{*}{0.1477} & \multirow{1}{*}{0.5960} & \multirow{1}{*}{0.5943} & \multirow{1}{*}{0.5480}  & \multirow{1}{*}{0.5970} & \multirow{1}{*}{\textbf{0.9783}}\\

         &&\multirow{1}{*}{MCC} & \multirow{1}{*}{0.1642} & \multirow{1}{*}{-0.6218} & \multirow{1}{*}{0.1720} & \multirow{1}{*}{0.1670} & \multirow{1}{*}{-0.1120}& \multirow{1}{*}{0.1730} & \multirow{1}{*}{\textbf{0.9297}}\\

         \multirow{2}{*}{\makecell{Entity\\Extraction}} & \multirow{2}{*}{NER} & \multirow{2}{*}{Entity-F1} & \multirow{2}{*}{0.1800}  & \multirow{2}{*}{0.6200} & \multirow{2}{*}{0.2900} & \multirow{2}{*}{0.2973} & \multirow{2}{*}{0.0231}  & \multirow{2}{*}{0.2910} & \multirow{2}{*}{\textbf{0.6993}}\\
         \\

         \multirow{2}{*}{\makecell{Relation\\Extraction}} & \multirow{2}{*}{FinRE} & \multirow{2}{*}{Relation-F1} & \multirow{2}{*}{0.1613} & \multirow{2}{*}{0.0054} & \multirow{2}{*}{0.1100} & \multirow{2}{*}{0.0540}  & \multirow{2}{*}{0.0100} & \multirow{2}{*}{0.1110} & \multirow{2}{*}{\textbf{0.5331}}\\
         \\
         \multirow{2}{*}{\makecell{Multiple\\Choice}} & \multirow{2}{*}{CFA} & \multirow{1}{*}{Weighted-F1} & \multirow{1}{*}{\textbf{0.7700}}  & \multirow{1}{*}{0.2200}  & \multirow{1}{*}{0.6720} & \multirow{1}{*}{0.5800} & \multirow{1}{*}{0.3993}  & \multirow{1}{*}{0.6730} & \multirow{1}{*}{0.7497}\\

         &&ACC & \multirow{1}{*}{\textbf{0.7700}}  & \multirow{1}{*}{0.2400}  & \multirow{1}{*}{0.6720} & \multirow{1}{*}{0.5800} & \multirow{1}{*}{0.3800}  & \multirow{1}{*}{0.6730} & \multirow{1}{*}{0.7500}\\
         
         \multirow{4}{*}{\makecell{Summarization}} & \multirow{4}{*}{EDTSUM} & \multirow{1}{*}{Rouge-1} & \multirow{1}{*}{0.1675}  & \multirow{1}{*}{0.1566} & \multirow{1}{*}{0.1480} & \multirow{1}{*}{0.1467} & \multirow{1}{*}{0.0622} & \multirow{1}{*}{0.1490} & \multirow{1}{*}{\textbf{0.5254}}\\

         &&\multirow{1}{*}{Rouge-2} & \multirow{1}{*}{0.0556}  & \multirow{1}{*}{0.0491} & \multirow{1}{*}{0.0440} & \multirow{1}{*}{0.0429} & \multirow{1}{*}{0.0085} & \multirow{1}{*}{0.0450} & \multirow{1}{*}{\textbf{0.3446}}\\

         &&\multirow{1}{*}{Rouge-L} & \multirow{1}{*}{0.1069}  & \multirow{1}{*}{0.1060} & \multirow{1}{*}{0.0860} & \multirow{1}{*}{0.0930} & \multirow{1}{*}{0.0412}  & \multirow{1}{*}{0.0870} & \multirow{1}{*}{\textbf{0.4705}}\\

         &&\multirow{1}{*}{BLEU} & \multirow{1}{*}{0.1192}  & \multirow{1}{*}{0.1361} & \multirow{1}{*}{0.1010} & \multirow{1}{*}{0.1085} & \multirow{1}{*}{0.0592} & \multirow{1}{*}{0.1020} & \multirow{1}{*}{\textbf{0.4512}}\\

         \multirow{2}{*}{\makecell{Question\\Answering}} & \multirow{1}{*}{Finqa} & \multirow{1}{*}{RMACC} & \multirow{1}{*}{0.1037}  & \multirow{1}{*}{0.0497}  & \multirow{1}{*}{0.0280} & \multirow{1}{*}{0.0470} & \multirow{1}{*}{0.0110} & \multirow{1}{*}{0.0290} & \multirow{1}{*}{\textbf{0.2258}}\\

         & \multirow{1}{*}{Convfinqa} & \multirow{1}{*}{RMACC} & \multirow{1}{*}{0.2540}  & \multirow{1}{*}{0.0953}  & \multirow{1}{*}{0.0650} & \multirow{1}{*}{0.1477} & \multirow{1}{*}{0.0772}  & \multirow{1}{*}{0.0660} & \multirow{1}{*}{\textbf{0.5053}}\\         
         \multirow{2}{*}{\makecell{Stock Movement\\Prediction}} 


         & \multirow{2}{*}{CMIN-US} & \multirow{1}{*}{Weighted-F1} & \multirow{1}{*}{0.5025} & \multirow{1}{*}{0.2639}  & \multirow{1}{*}{0.4130} & \multirow{1}{*}{0.3722} & \multirow{1}{*}{0.3379} & \multirow{1}{*}{0.4140} & \multirow{1}{*}{\textbf{0.5036}}\\

         && \multirow{1}{*}{ACC} & \multirow{1}{*}{\textbf{0.5149}} & \multirow{1}{*}{0.3446}  & \multirow{1}{*}{0.5120} & \multirow{1}{*}{0.4955} & \multirow{1}{*}{0.4154} & \multirow{1}{*}{0.5130} & \multirow{1}{*}{0.5144}\\
         
        \midrule
    \end{tabular}
    \label{tab:main_results_en}
\end{table*}


\begin{table*}
    \centering
    \caption{Performance metrics of financial large language models across chinese tasks like Sentiment Analysis, Classification, and Summarization. Models include GPT-4o, Llama-3-8B, Qwen-3-8B, CFGPT-7B, DISC-FinLLM, and Touchstone-GPT. The best results of each dataset are marked in \textbf{bold}.}
    \renewcommand\arraystretch{1}
    \scriptsize
    \begin{tabular}{c@{}c@{}c@{}c@{}c@{}c@{}c@{}c@{}c@{}c@{}c@{}c}
        \toprule
        \multirow{1}{*}{\textbf{Task}} & \multicolumn{1}{c}{\textbf{Dataset}} & \multicolumn{1}{c}{\textbf{Metrics}} & \multicolumn{1}{c}{\textbf{\makecell{GPT-4o}}} & \multicolumn{1}{c}{\textbf{\makecell{Qwen-3-8B\\Instruct}}} & \multicolumn{1}{c}{\textbf{\makecell{Llama-3-8B\\Instruct}}} & \multicolumn{1}{c}{\textbf{\makecell{CFGPT1-7B\\Full}}} & \multicolumn{1}{c}{\textbf{\makecell{DISC-FinLLM\\Full}}}  & \multicolumn{1}{c}{\textbf{\makecell{Touchstone\\GPT}}}\\
         \midrule
         \multirow{2}{*}{\makecell{Sentiment\\Analysis}} & \multirow{2}{*}{FinFe-CN} & \multirow{1}{*}{Weighted-F1} & \multirow{1}{*}{0.6593}  & \multirow{1}{*}{0.6478} & \multirow{1}{*}{0.3633} &\multirow{1}{*}{0.2528} &\multirow{1}{*}{0.4177} &\multirow{1}{*}{\textbf{0.7888}}\\

         &&ACC & \multirow{1}{*}{0.6500}  & \multirow{1}{*}{0.6632} & \multirow{1}{*}{0.4891} & \multirow{1}{*}{0.2732} & \multirow{1}{*}{0.4292}  &\multirow{1}{*}{\textbf{0.7936}}\\

         \multirow{2}{*}{\makecell{Classification}} & \multirow{2}{*}{FinNL-CN} & \multirow{2}{*}{ORMACC}  & \multirow{2}{*}{0.3303}   & \multirow{2}{*}{0.0824} & \multirow{2}{*}{0.0747} & \multirow{2}{*}{0.0894} & \multirow{2}{*}{0.0011} & \multirow{2}{*}{\textbf{0.8360}}\\
         \\

         \multirow{2}{*}{\makecell{Entity\\Extraction}} & \multirow{2}{*}{FinESE-CN} & \multirow{2}{*}{ORMACC}  & \multirow{2}{*}{0.6867}   & \multirow{2}{*}{0.3879} & \multirow{2}{*}{0.3088} & \multirow{2}{*}{0.3863} & \multirow{2}{*}{0.4346}  &\multirow{2}{*}{\textbf{0.9074}}\\
         \\

        \multirow{2}{*}{\makecell{Relation\\Extraction}} & \multirow{2}{*}{FinRE-CN} & \multirow{2}{*}{RMACC}  & \multirow{2}{*}{0.2754}   & \multirow{2}{*}{0.1532} & \multirow{2}{*}{0.1296} & \multirow{2}{*}{0.0678} & \multirow{2}{*}{0.1182}& \multirow{2}{*}{\textbf{0.6541}}\\
         \\
         
         \multirow{4}{*}{\makecell{Multiple\\Choice}} & \multirow{2}{*}{FinEval} & \multirow{1}{*}{Weighted-F1} & \multirow{1}{*}{\textbf{0.7364}}  & \multirow{1}{*}{0.7332} & \multirow{1}{*}{0.4432} & \multirow{1}{*}{0.3543} & \multirow{1}{*}{0.4288} & \multirow{1}{*}{0.7361}\\
         
         &&\multirow{1}{*}{ACC} & \multirow{1}{*}{\textbf{0.7353}}  & \multirow{1}{*}{0.7337} & \multirow{1}{*}{0.4471} & \multirow{1}{*}{0.3529} & \multirow{1}{*}{0.4294}  & \multirow{1}{*}{\textbf{0.7353}}\\

         & \multirow{2}{*}{CPA} & \multirow{1}{*}{Weighted-F1} & \multirow{1}{*}{0.6312}  & \multirow{1}{*}{0.7159} & \multirow{1}{*}{0.3421} & \multirow{1}{*}{0.3543} & \multirow{1}{*}{0.3451} & \multirow{1}{*}{\textbf{0.9238}}\\
         
         &&\multirow{1}{*}{ACC} & \multirow{1}{*}{0.6309}  & \multirow{1}{*}{0.7162} & \multirow{1}{*}{0.3504} & \multirow{1}{*}{0.3553} & \multirow{1}{*}{0.3518} & \multirow{1}{*}{\textbf{0.9238}}\\
         
         \multirow{4}{*}{\makecell{Summarization}} & \multirow{4}{*}{FinNA-CN} & \multirow{1}{*}{Rouge-1}  & \multirow{1}{*}{0.3197}   & \multirow{1}{*}{0.3528} & \multirow{1}{*}{0.3477} & \multirow{1}{*}{0.1018} & \multirow{1}{*}{0.3486}  & \multirow{1}{*}{\textbf{0.5526}}\\

         &&\multirow{1}{*}{Rouge-2}  & \multirow{1}{*}{0.1434}  & \multirow{1}{*}{0.1699} & \multirow{1}{*}{0.1702} & \multirow{1}{*}{0.0263} & \multirow{1}{*}{0.1678}  & \multirow{1}{*}{\textbf{0.3603}}\\

         &&\multirow{1}{*}{Rouge-L}  & \multirow{1}{*}{0.2511}  & \multirow{1}{*}{0.2846} & \multirow{1}{*}{0.2802} & \multirow{1}{*}{0.0650} & \multirow{1}{*}{0.2997} & \multirow{1}{*}{\textbf{0.5214}}\\

         &&\multirow{1}{*}{BLEU}  & \multirow{1}{*}{0.1423}   & \multirow{1}{*}{0.1743} & \multirow{1}{*}{0.1672} & \multirow{1}{*}{0.0238} & \multirow{1}{*}{0.1885}  & \multirow{1}{*}{\textbf{0.3944}}\\

        \multirow{2}{*}{\makecell{Question\\Answering}} & \multirow{1}{*}{FinQa-CN} & \multirow{1}{*}{RMACC}  & \multirow{1}{*}{0.6578}  & \multirow{1}{*}{0.5245} & \multirow{1}{*}{0.4540} & \multirow{1}{*}{0.1126} & \multirow{1}{*}{0.3949}  &\multirow{1}{*}{\textbf{0.9214}}\\

         &\multirow{1}{*}{FinCQa-CN} & \multirow{1}{*}{RMACC}  & \multirow{1}{*}{0.4765}  & \multirow{1}{*}{0.3624} & \multirow{1}{*}{0.3787} & \multirow{1}{*}{0.2714} & \multirow{1}{*}{0.2134}  &\multirow{1}{*}{\textbf{0.8552}}\\
         
         \multirow{2}{*}{\makecell{Stock Movement\\Prediction}} 
         & \multirow{2}{*}{CMIN-CN} & \multirow{1}{*}{Weighted-F1} & \multirow{1}{*}{\textbf{0.4858}}  & \multirow{1}{*}{0.4165} & \multirow{1}{*}{0.4497} & \multirow{1}{*}{0.3549} & \multirow{1}{*}{0.0329} & \multirow{1}{*}{0.4735}\\

         && \multirow{1}{*}{ACC} & \multirow{1}{*}{0.4988}   & \multirow{1}{*}{0.4825} & \multirow{1}{*}{0.4858} & \multirow{1}{*}{0.3584} & \multirow{1}{*}{0.0332} & \multirow{1}{*}{\textbf{0.4878}}\\
         
        \midrule
    \end{tabular}
    \label{tab:main_results_cn}
\end{table*}


\section{Related Works}

\subsection{Financial Large Language Models}
In recent years, large language models (LLMs) tailored for the financial domain have gained significant attention. BloombertGPT \citep{wu2023bloomberggpt} marked the beginning of the FinLLM era. FinGPT \citep{yang2023fingpt} introduced an open-source framework emphasizing a data-centric approach with lightweight low-rank adaptation techniques. PIXIU \citep{xie2023pixiu} provided a comprehensive framework, presenting the first financial LLM fine-tuned on LLaMA with a 136K instruction dataset and evaluation benchmark. CFGPT \citep{li2023cfgpt} developed a Chinese Financial Generative Pre-trained Transformer framework, encompassing dataset, model, and deployment capabilities. DISC-FinLLM \citep{chen2023disc} enhanced general LLMs through a multiple experts fine-tuning framework, expanding domain-specific capabilities.

\subsection{Benchmarks for FinLLMs}
In the current landscape of open-source financial evaluation benchmarks, FLUE \citep{shah2022flue} was the first open-source benchmark introduced for financial language understanding in English. It includes five financial tasks: Financial Sentiment Analysis \citep{malo2014fpb,maia2018finqa_sa}, Financial News Classification \citep{sinha2021headlines}, Structure Boundary Detection \citep{finsbd3}, Named Entity Recognition \citep{alvarado2015NER}, and Question Answering \citep{FiQA_flue}. Building on FLUE, FinGPT \citep{yang2023fingpt} introduced the Financial Relation Extraction task with the addition of the FinRED dataset \citep{sharma2022finred}, while also expanding the sentiment analysis task by incorporating TFNS \citep{tfns2022} and NWGI \citep{nwgi2023} datasets. Additionally, FinGPT created prompt-based instructions across all datasets, culminating in the Financial Instruction Tuning Benchmark. Around the same time, other open-source benchmarks, such as PIXIU \citep{xie2023pixiu} and FinBen \citep{xie2024finben}, were developed. PIXIU includes a wide range of financial tasks and datasets, while FinBen, expanding on PIXIU, introduced numerous supplementary financial datasets to evaluate LLMs comprehensively and gauge their proficiency across various financial scenarios.In the Chinese domain, BBT-Benchmark \citep{lu2023bbt} was the first Chinese financial benchmark, covering multiple financial NLP tasks, including sentiment analysis, news classification, summarization, relation extraction, and question answering. Later, FinEval \citep{zhang2023fineval} and FinanceIQ \citep{zhang2023xuanyuan}benchmark emerged, providing an evaluation framework for LLMs in financial contexts, encompassing multiple-choice questions across finance, economics, accounting, and certification. DISC-FinLLM \citep{chen2023disc} evaluated financial LLMs on top of these benchmarks, introducing additional tasks focused on financial calculations and current events to assess LLMs in applied financial scenarios. The most recent Chinese Financial benchmark, CFBenchmark \citep{lei2023cfbenchmark}, introduced three key tasks like Financial Entity Recognition, expanding evaluation resources for Chinese financial tasks.

The increasing number of benchmarks in both English and Chinese financial domains has established valuable evaluation frameworks for LLMs in finance, though challenges remain. For example, data quality across various English financial datasets is inconsistent, limiting LLMs from reaching their full potential, as seen in FinBen and PIXIU, where low-quality datasets yielded suboptimal performance metrics. In Chinese benchmarks, tasks are often skewed toward multiple-choice questions, with limited alignment with English financial tasks. Moreover, no bilingual benchmark currently exists for comprehensive LLM evaluation in financial NLP tasks across English and Chinese. In this work, we integrate representative datasets from both languages, creating FinLLM-benchmark. We compile and align eight financial NLP tasks in English and Chinese, selecting high-quality datasets for each task and appropriate metrics to effectively assess model performance.


\section{Conclusion}
In this study, we introduce the Golden Touchstone benchmark, the inaugural structured and comprehensive bilingual benchmark specifically designed for English-Chinese financial NLP. This benchmark encompasses a wide array of financial NLP tasks, including Natural Language Understanding (NLU) and Natural Language Generation (NLG) across eight categories: Sentiment Analysis, Classification, Entity Extraction, Summarization, Stock Market Prediction, Question Answering, Relation Extraction, and Multiple Choice. By leveraging existing high-quality open-source financial datasets, we curated representative datasets and selected appropriate evaluation metrics for each task category. Utilizing these resources, we conducted extensive evaluations of current models such as GPT-4o and prominent open-source financial LLMs, including FinGPT and FinMA, thereby establishing performance benchmarks for financial LLMs within bilingual contexts. Moreover, we contributed to the community by open-sourcing Touchstone-GPT, a robust financial LLM that employs a two-stage training approach and has demonstrated superior input-based inference capabilities on the Golden Touchstone benchmark compared to GPT-4o.
Our open-source initiative provides a bilingual English-Chinese evaluation framework aimed at fostering the sustainable development of LLMs in a multilingual financial environment.

\section{Limitations}
Despite these advancements, the benchmark currently exhibits certain limitations, including a limited range of NLG tasks and a focus solely on single-modality. Future enhancements will include the integration of additional NLG tasks, such as extended text generation for financial report analysis and more sophisticated sentiment assessments. Dataset biases and model scalability are also challenges.
Furthermore, we plan to expand the benchmark to cover other financial sectors such as insurance and futures trading, thus broadening the scope and applicability of financial LLM assessments across diverse scenarios. 
Also, the performance of Touchstone-GPT on specific tasks within the Golden Touchstone benchmark, particularly in stock market prediction, requires further improvement.
Our subsequent research will explore the incorporation of agent-based and retrieval-augmented generation (RAG) methods to augment the model’s capabilities in numerical computation and real-time news analysis. 
Additionally, we aim to venture into multimodal modeling, integrating visual data and time-series data for tasks such as financial time-series forecasting, financial chart analysis, and content generation.

\section{Ethical Statements}
We do not see our work to have possible harmful outcomes. We follow the ACL ethical guidelines when conducting the research in this paper.

\clearpage
\bibliography{anthology,custom}

@article{hu2024no,
  title={No language is an island: Unifying chinese and english in financial large language models, instruction data, and benchmarks},
  author={Hu, Gang and Qin, Ke and Yuan, Chenhan and Peng, Min and Lopez-Lira, Alejandro and Wang, Benyou and Ananiadou, Sophia and Huang, Jimin and Xie, Qianqian},
  journal={arXiv preprint arXiv:2403.06249},
  year={2024}
}

@article{nie2024cfinbench,
  title={CFinBench: A Comprehensive Chinese Financial Benchmark for Large Language Models},
  author={Nie, Ying and Yan, Binwei and Guo, Tianyu and Liu, Hao and Wang, Haoyu and He, Wei and Zheng, Binfan and Wang, Weihao and Li, Qiang and Sun, Weijian and others},
  journal={arXiv preprint arXiv:2407.02301},
  year={2024}
}

@article{liu2024deepseekv3,
  title={Deepseek-v3 technical report},
  author={Liu, Aixin and Feng, Bei and Xue, Bing and Wang, Bingxuan and Wu, Bochao and Lu, Chengda and Zhao, Chenggang and Deng, Chengqi and Zhang, Chenyu and Ruan, Chong and others},
  journal={arXiv preprint arXiv:2412.19437},
  year={2024}
}

@article{yang2025qwen3,
  title={Qwen3 Technical Report},
  author={Yang, An and Li, Anfeng and Yang, Baosong and Zhang, Beichen and Hui, Binyuan and Zheng, Bo and Yu, Bowen and Gao, Chang and Huang, Chengen and Lv, Chenxu and others},
  journal={arXiv preprint arXiv:2505.09388},
  year={2025}
}

@article{zhang2022fengshenbang,
  title={Fengshenbang 1.0: Being the foundation of chinese cognitive intelligence},
  author={Zhang, Jiaxing and Gan, Ruyi and Wang, Junjie and Zhang, Yuxiang and Zhang, Lin and Yang, Ping and Gao, Xinyu and Wu, Ziwei and Dong, Xiaoqun and He, Junqing and others},
  journal={arXiv preprint arXiv:2209.02970},
  year={2022}
}

@article{gan2023ziya2,
  title={Ziya2: Data-centric Learning is All LLMs Need},
  author={Gan, Ruyi and Wu, Ziwei and Sun, Renliang and Lu, Junyu and Wu, Xiaojun and Zhang, Dixiang and Pan, Kunhao and Yang, Ping and Yang, Qi and Zhang, Jiaxing and others},
  journal={arXiv preprint arXiv:2311.03301},
  year={2023}
}

@misc{glm2024chatglm,
      title={ChatGLM: A Family of Large Language Models from GLM-130B to GLM-4 All Tools},
      author={Team GLM and Aohan Zeng and Bin Xu and Bowen Wang and Chenhui Zhang and Da Yin and Diego Rojas and Guanyu Feng and Hanlin Zhao and Hanyu Lai and Hao Yu and Hongning Wang and Jiadai Sun and Jiajie Zhang and Jiale Cheng and Jiayi Gui and Jie Tang and Jing Zhang and Juanzi Li and Lei Zhao and Lindong Wu and Lucen Zhong and Mingdao Liu and Minlie Huang and Peng Zhang and Qinkai Zheng and Rui Lu and Shuaiqi Duan and Shudan Zhang and Shulin Cao and Shuxun Yang and Weng Lam Tam and Wenyi Zhao and Xiao Liu and Xiao Xia and Xiaohan Zhang and Xiaotao Gu and Xin Lv and Xinghan Liu and Xinyi Liu and Xinyue Yang and Xixuan Song and Xunkai Zhang and Yifan An and Yifan Xu and Yilin Niu and Yuantao Yang and Yueyan Li and Yushi Bai and Yuxiao Dong and Zehan Qi and Zhaoyu Wang and Zhen Yang and Zhengxiao Du and Zhenyu Hou and Zihan Wang},
      year={2024},
      eprint={2406.12793},
      archivePrefix={arXiv},
      primaryClass={id='cs.CL' full_name='Computation and Language' is_active=True alt_name='cmp-lg' in_archive='cs' is_general=False description='Covers natural language processing. Roughly includes material in ACM Subject Class I.2.7. Note that work on artificial languages (programming languages, logics, formal systems) that does not explicitly address natural-language issues broadly construed (natural-language processing, computational linguistics, speech, text retrieval, etc.) is not appropriate for this area.'}
}

@article{wu2023bloomberggpt,
  title={Bloomberggpt: A large language model for finance},
  author={Wu, Shijie and Irsoy, Ozan and Lu, Steven and Dabravolski, Vadim and Dredze, Mark and Gehrmann, Sebastian and Kambadur, Prabhanjan and Rosenberg, David and Mann, Gideon},
  journal={arXiv preprint arXiv:2303.17564},
  year={2023}
}

@article{lopez2023chatgpt_stock_move,
  title={Can chatgpt forecast stock price movements? return predictability and large language models},
  author={Lopez-Lira, Alejandro and Tang, Yuehua},
  journal={arXiv preprint arXiv:2304.07619},
  year={2023}
}

@article{yang2023fingpt,
  title={Fingpt: Open-source financial large language models},
  author={Yang, Hongyang and Liu, Xiao-Yang and Wang, Christina Dan},
  journal={arXiv preprint arXiv:2306.06031},
  year={2023}
}

@article{wang2023fingptbench,
  title={Fingpt: Instruction tuning benchmark for open-source large language models in financial datasets},
  author={Wang, Neng and Yang, Hongyang and Wang, Christina Dan},
  journal={arXiv preprint arXiv:2310.04793},
  year={2023}
}

@article{xie2023pixiu,
  title={Pixiu: A large language model, instruction data and evaluation benchmark for finance},
  author={Xie, Qianqian and Han, Weiguang and Zhang, Xiao and Lai, Yanzhao and Peng, Min and Lopez-Lira, Alejandro and Huang, Jimin},
  journal={arXiv preprint arXiv:2306.05443},
  year={2023}
}

@article{xie2024finben,
  title={The finben: An holistic financial benchmark for large language models},
  author={Xie, Qianqian and Han, Weiguang and Chen, Zhengyu and Xiang, Ruoyu and Zhang, Xiao and He, Yueru and Xiao, Mengxi and Li, Dong and Dai, Yongfu and Feng, Duanyu and others},
  journal={arXiv preprint arXiv:2402.12659},
  year={2024}
}

@article{li2023cfgpt,
  title={CFGPT: Chinese financial assistant with large language model},
  author={Li, Jiangtong and Bian, Yuxuan and Wang, Guoxuan and Lei, Yang and Cheng, Dawei and Ding, Zhijun and Jiang, Changjun},
  journal={arXiv preprint arXiv:2309.10654},
  year={2023}
}

@article{chen2023disc,
  title={DISC-FinLLM: A Chinese financial large language model based on multiple experts fine-tuning},
  author={Chen, Wei and Wang, Qiushi and Long, Zefei and Zhang, Xianyin and Lu, Zhongtian and Li, Bingxuan and Wang, Siyuan and Xu, Jiarong and Bai, Xiang and Huang, Xuanjing and others},
  journal={arXiv preprint arXiv:2310.15205},
  year={2023}
}

@inproceedings{zhang2023xuanyuan,
  title={Xuanyuan 2.0: A large chinese financial chat model with hundreds of billions parameters},
  author={Zhang, Xuanyu and Yang, Qing},
  booktitle={Proceedings of the 32nd ACM international conference on information and knowledge management},
  pages={4435--4439},
  year={2023}
}

@article{yang2024qwen2,
  title={Qwen2 Technical Report},
  author={Yang, An and Yang, Baosong and Hui, Binyuan and Zheng, Bo and Yu, Bowen and Zhou, Chang and Li, Chengpeng and Li, Chengyuan and Liu, Dayiheng and Huang, Fei and others},
  journal={arXiv preprint arXiv:2407.10671},
  year={2024}
}

@misc{deepseekv2,
      title={DeepSeek-V2: A Strong, Economical, and Efficient Mixture-of-Experts Language Model}, 
      author={DeepSeek-AI},
      year={2024},
      eprint={2405.04434},
      archivePrefix={arXiv},
      primaryClass={cs.CL}
}

@article{anthropic2024claude,
  title={The claude 3 model family: Opus, sonnet, haiku},
  author={Anthropic, AI},
  journal={Claude-3 Model Card},
  year={2024}
}

@article{young2024yi,
  title={Yi: Open foundation models by 01. ai},
  author={Young, Alex and Chen, Bei and Li, Chao and Huang, Chengen and Zhang, Ge and Zhang, Guanwei and Li, Heng and Zhu, Jiangcheng and Chen, Jianqun and Chang, Jing and others},
  journal={arXiv preprint arXiv:2403.04652},
  year={2024}
}

@article{llama3,
title={Llama 3 Model Card},
author={AI@Meta},
year={2024},
url = {https://github.com/meta-llama/llama3/blob/main/MODEL_CARD.md}
}

@article{jiang2024mixtral,
  title={Mixtral of experts},
  author={Jiang, Albert Q and Sablayrolles, Alexandre and Roux, Antoine and Mensch, Arthur and Savary, Blanche and Bamford, Chris and Chaplot, Devendra Singh and Casas, Diego de las and Hanna, Emma Bou and Bressand, Florian and others},
  journal={arXiv preprint arXiv:2401.04088},
  year={2024}
}

@article{team2023gemini,
  title={Gemini: a family of highly capable multimodal models},
  author={Team, Gemini and Anil, Rohan and Borgeaud, Sebastian and Wu, Yonghui and Alayrac, Jean-Baptiste and Yu, Jiahui and Soricut, Radu and Schalkwyk, Johan and Dai, Andrew M and Hauth, Anja and others},
  journal={arXiv preprint arXiv:2312.11805},
  year={2023}
}

@article{openai2023gpt4,
  title={GPT-4 technical report},
  author={OpenAI, R},
  journal={arXiv},
  pages={2303--08774},
  year={2023}
}

@article{touvron2023llama,
  title={Llama: Open and efficient foundation language models},
  author={Touvron, Hugo and Lavril, Thibaut and Izacard, Gautier and Martinet, Xavier and Lachaux, Marie-Anne and Lacroix, Timoth{\'e}e and Rozi{\`e}re, Baptiste and Goyal, Naman and Hambro, Eric and Azhar, Faisal and others},
  journal={arXiv preprint arXiv:2302.13971},
  year={2023}
}

@article{touvron2023llama2,
  title={Llama 2: Open foundation and fine-tuned chat models},
  author={Touvron, Hugo and Martin, Louis and Stone, Kevin and Albert, Peter and Almahairi, Amjad and Babaei, Yasmine and Bashlykov, Nikolay and Batra, Soumya and Bhargava, Prajjwal and Bhosale, Shruti and others},
  journal={arXiv preprint arXiv:2307.09288},
  year={2023}
}

@article{bai2023qwen,
  title={Qwen technical report},
  author={Bai, Jinze and Bai, Shuai and Chu, Yunfei and Cui, Zeyu and Dang, Kai and Deng, Xiaodong and Fan, Yang and Ge, Wenbin and Han, Yu and Huang, Fei and others},
  journal={arXiv preprint arXiv:2309.16609},
  year={2023}
}

@inproceedings{zeng2023glm-130b,
  title={{GLM}-130B: An Open Bilingual Pre-trained Model},
  author={Aohan Zeng and Xiao Liu and Zhengxiao Du and Zihan Wang and Hanyu Lai and Ming Ding and Zhuoyi Yang and Yifan Xu and Wendi Zheng and Xiao Xia and Weng Lam Tam and Zixuan Ma and Yufei Xue and Jidong Zhai and Wenguang Chen and Zhiyuan Liu and Peng Zhang and Yuxiao Dong and Jie Tang},
  booktitle={The Eleventh International Conference on Learning Representations (ICLR)},
  year={2023},
  url={https://openreview.net/forum?id=-Aw0rrrPUF}
}

@article{baichuan2023baichuan2,
  title={Baichuan 2: Open Large-scale Language Models},
  author={Baichuan},
  journal={arXiv preprint arXiv:2309.10305},
  url={https://arxiv.org/abs/2309.10305},
  year={2023}
}

@article{lu2023bbt,
  title={Bbt-fin: Comprehensive construction of chinese financial domain pre-trained language model, corpus and benchmark},
  author={Lu, Dakuan and Wu, Hengkui and Liang, Jiaqing and Xu, Yipei and He, Qianyu and Geng, Yipeng and Han, Mengkun and Xin, Yingsi and Xiao, Yanghua},
  journal={arXiv preprint arXiv:2302.09432},
  year={2023}
}

@article{lei2023cfbenchmark,
  title={CFBenchmark: Chinese financial assistant benchmark for large language model},
  author={Lei, Yang and Li, Jiangtong and Jiang, Ming and Hu, Junjie and Cheng, Dawei and Ding, Zhijun and Jiang, Changjun},
  journal={arXiv preprint arXiv:2311.05812},
  year={2023}
}

@article{zhang2023fineval,
  title={Fineval: A chinese financial domain knowledge evaluation benchmark for large language models},
  author={Zhang, Liwen and Cai, Weige and Liu, Zhaowei and Yang, Zhi and Dai, Wei and Liao, Yujie and Qin, Qianru and Li, Yifei and Liu, Xingyu and Liu, Zhiqiang and others},
  journal={arXiv preprint arXiv:2308.09975},
  year={2023}
}

@article{ouyang2022instruct_gpt,
  title={Training language models to follow instructions with human feedback},
  author={Ouyang, Long and Wu, Jeffrey and Jiang, Xu and Almeida, Diogo and Wainwright, Carroll and Mishkin, Pamela and Zhang, Chong and Agarwal, Sandhini and Slama, Katarina and Ray, Alex and others},
  journal={Advances in Neural Information Processing Systems},
  volume={35},
  pages={27730--27744},
  year={2022}
}

@article{shah2022flue,
  title={When flue meets flang: Benchmarks and large pre-trained language model for financial domain},
  author={Shah, Raj Sanjay and Chawla, Kunal and Eidnani, Dheeraj and Shah, Agam and Du, Wendi and Chava, Sudheer and Raman, Natraj and Smiley, Charese and Chen, Jiaao and Yang, Diyi},
  journal={arXiv preprint arXiv:2211.00083},
  year={2022}
}

@article{brown2020gpt3,
  title={Language models are few-shot learners},
  author={Brown, Tom and Mann, Benjamin and Ryder, Nick and Subbiah, Melanie and Kaplan, Jared D and Dhariwal, Prafulla and Neelakantan, Arvind and Shyam, Pranav and Sastry, Girish and Askell, Amanda and others},
  journal={Advances in neural information processing systems},
  volume={33},
  pages={1877--1901},
  year={2020}
}

@article{cui2023chatlaw,
  title={Chatlaw: Open-source legal large language model with integrated external knowledge bases},
  author={Cui, Jiaxi and Li, Zongjian and Yan, Yang and Chen, Bohua and Yuan, Li},
  journal={arXiv preprint arXiv:2306.16092},
  year={2023}
}

@article{xiao2021lawformer,
  title={Lawformer: A pre-trained language model for chinese legal long documents},
  author={Xiao, Chaojun and Hu, Xueyu and Liu, Zhiyuan and Tu, Cunchao and Sun, Maosong},
  journal={AI Open},
  volume={2},
  pages={79--84},
  year={2021},
  publisher={Elsevier}
}

@article{thirunavukarasu2023llm_med,
  title={Large language models in medicine},
  author={Thirunavukarasu, Arun James and Ting, Darren Shu Jeng and Elangovan, Kabilan and Gutierrez, Laura and Tan, Ting Fang and Ting, Daniel Shu Wei},
  journal={Nature medicine},
  volume={29},
  number={8},
  pages={1930--1940},
  year={2023},
  publisher={Nature Publishing Group US New York}
}

@article{tian2023chimed,
  title={Chimed-gpt: A chinese medical large language model with full training regime and better alignment to human preferences},
  author={Tian, Yuanhe and Gan, Ruyi and Song, Yan and Zhang, Jiaxing and Zhang, Yongdong},
  journal={arXiv preprint arXiv:2311.06025},
  year={2023}
}

@article{malo2014fpb,
  title={Good debt or bad debt: Detecting semantic orientations in economic texts},
  author={Malo, Pekka and Sinha, Ankur and Korhonen, Pekka and Wallenius, Jyrki and Takala, Pyry},
  journal={Journal of the Association for Information Science and Technology},
  volume={65},
  number={4},
  pages={782--796},
  year={2014},
  publisher={Wiley Online Library}
}

@inproceedings{maia2018finqa_sa,
author = {Maia, Macedo and Handschuh, Siegfried and Freitas, Andre and Davis, Brian and McDermott, Ross and Zarrouk, Manel and Balahur, Alexandra},
year = {2018},
month = {04},
pages = {1941-1942},
title = {WWW'18 Open Challenge: Financial Opinion Mining and Question Answering},
isbn = {9781450356404},
journal = {WWW '18: Companion Proceedings of the The Web Conference 2018},
doi = {10.1145/3184558.3192301}
}

@inproceedings{sinha2021headlines,
  title={Impact of news on the commodity market: Dataset and results},
  author={Sinha, Ankur and Khandait, Tanmay},
  booktitle={Advances in Information and Communication: Proceedings of the 2021 Future of Information and Communication Conference (FICC), Volume 2},
  pages={589--601},
  year={2021},
  organization={Springer}
}

@inproceedings{alvarado2015NER,
  title={Domain adaption of named entity recognition to support credit risk assessment},
  author={Alvarado, Julio Cesar Salinas and Verspoor, Karin and Baldwin, Timothy},
  booktitle={Proceedings of the Australasian Language Technology Association Workshop 2015},
  pages={84--90},
  year={2015}
}

@inproceedings{sharma2022finred,
  title={FinRED: A dataset for relation extraction in financial domain},
  author={Sharma, Soumya and Nayak, Tapas and Bose, Arusarka and Meena, Ajay Kumar and Dasgupta, Koustuv and Ganguly, Niloy and Goyal, Pawan},
  booktitle={Companion Proceedings of the Web Conference 2022},
  pages={595--597},
  year={2022}
}

@misc{finsbd3,
  title={Financial SBD 3},
  author={FinSBD3},
  year={2021},
  howpublished={\url{https://sites.google.com/nlg.csie.ntu.edu.tw/finweb2021/shared-task-finsbd-3}}
}

@inproceedings{FiQA_flue,
author = {Maia, Macedo and Handschuh, Siegfried and Freitas, Andr\'{e} and Davis, Brian and McDermott, Ross and Zarrouk, Manel and Balahur, Alexandra},
title = {WWW'18 Open Challenge: Financial Opinion Mining and Question Answering},
year = {2018},
isbn = {9781450356404},
publisher = {International World Wide Web Conferences Steering Committee},
address = {Republic and Canton of Geneva, CHE},
url = {https://doi.org/10.1145/3184558.3192301},
doi = {10.1145/3184558.3192301},
booktitle = {Companion Proceedings of the The Web Conference 2018},
pages = {1941–1942},
numpages = {2},
location = {Lyon, France},
series = {WWW '18}
}

@article{peng2023Alpaca_gpt4,
  title={Instruction tuning with gpt-4},
  author={Peng, Baolin and Li, Chunyuan and He, Pengcheng and Galley, Michel and Gao, Jianfeng},
  journal={arXiv preprint arXiv:2304.03277},
  year={2023}
}

@article{kingma2014adam,
  title={Adam: A Method for Stochastic Optimization},
  author={Kingma, Diederik P and Ba, Jimmy},
  journal={arXiv e-prints},
  pages={arXiv--1412},
  year={2014}
}

@inproceedings{rajbhandari2020zero,
  title={Zero: Memory optimizations toward training trillion parameter models},
  author={Rajbhandari, Samyam and Rasley, Jeff and Ruwase, Olatunji and He, Yuxiong},
  booktitle={SC20: International Conference for High Performance Computing, Networking, Storage and Analysis},
  pages={1--16},
  year={2020},
  organization={IEEE}
}

@inproceedings{he2016resnet,
  title={Deep residual learning for image recognition},
  author={He, Kaiming and Zhang, Xiangyu and Ren, Shaoqing and Sun, Jian},
  booktitle={Proceedings of the IEEE conference on computer vision and pattern recognition},
  pages={770--778},
  year={2016}
}

@article{goyal2017warmup,
  title={Accurate, large minibatch sgd: Training imagenet in 1 hour},
  author={Goyal, Priya and Doll{\'a}r, Piotr and Girshick, Ross and Noordhuis, Pieter and Wesolowski, Lukasz and Kyrola, Aapo and Tulloch, Andrew and Jia, Yangqing and He, Kaiming},
  journal={arXiv preprint arXiv:1706.02677},
  year={2017}
}

@inproceedings{szegedy2016cosine,
  title={Rethinking the inception architecture for computer vision},
  author={Szegedy, Christian and Vanhoucke, Vincent and Ioffe, Sergey and Shlens, Jon and Wojna, Zbigniew},
  booktitle={Proceedings of the IEEE conference on computer vision and pattern recognition},
  pages={2818--2826},
  year={2016}
}

@article{wang2019bf16,
  title={BFloat16: The secret to high performance on Cloud TPUs},
  author={Wang, Shibo and Kanwar, Pankaj},
  journal={Google Cloud Blog},
  volume={4},
  number={1},
  year={2019}
}

@article{micikevicius2017mixed,
  title={Mixed precision training},
  author={Micikevicius, Paulius and Narang, Sharan and Alben, Jonah and Diamos, Gregory and Elsen, Erich and Garcia, David and Ginsburg, Boris and Houston, Michael and Kuchaiev, Oleksii and Venkatesh, Ganesh and others},
  journal={arXiv preprint arXiv:1710.03740},
  year={2017}
}

@article{chen2016gradient_ckpt,
  title={Training deep nets with sublinear memory cost},
  author={Chen, Tianqi and Xu, Bing and Zhang, Chiyuan and Guestrin, Carlos},
  journal={arXiv preprint arXiv:1604.06174},
  year={2016}
}

@article{shoeybi2019megatron,
  title={Megatron-lm: Training multi-billion parameter language models using model parallelism},
  author={Shoeybi, Mohammad and Patwary, Mostofa and Puri, Raul and LeGresley, Patrick and Casper, Jared and Catanzaro, Bryan},
  journal={arXiv preprint arXiv:1909.08053},
  year={2019}
}

@inproceedings{zheng2024llamafactory,
  title={LlamaFactory: Unified Efficient Fine-Tuning of 100+ Language Models},
  author={Yaowei Zheng and Richong Zhang and Junhao Zhang and Yanhan Ye and Zheyan Luo and Zhangchi Feng and Yongqiang Ma},
  booktitle={Proceedings of the 62nd Annual Meeting of the Association for Computational Linguistics (Volume 3: System Demonstrations)},
  address={Bangkok, Thailand},
  publisher={Association for Computational Linguistics},
  year={2024},
  url={http://arxiv.org/abs/2403.13372}
}

@article{krell2021packing,
  title={Efficient Sequence Packing without Cross-contamination: Accelerating Large Language Models without Impacting Performance},
  author={Krell, Mario Michael and Kosec, Matej and Perez, Sergio P and Fitzgibbon, Andrew},
  journal={arXiv preprint arXiv:2107.02027},
  year={2021},
  url={http://arxiv.org/abs/2107.02027}
}

@inproceedings{wolf-etal-2020-transformers,
    title = "Transformers: State-of-the-Art Natural Language Processing",
    author = "Thomas Wolf and Lysandre Debut and Victor Sanh and Julien Chaumond and Clement Delangue and Anthony Moi and Pierric Cistac and Tim Rault and Rémi Louf and Morgan Funtowicz and Joe Davison and Sam Shleifer and Patrick von Platen and Clara Ma and Yacine Jernite and Julien Plu and Canwen Xu and Teven Le Scao and Sylvain Gugger and Mariama Drame and Quentin Lhoest and Alexander M. Rush",
    booktitle = "Proceedings of the 2020 Conference on Empirical Methods in Natural Language Processing: System Demonstrations",
    month = oct,
    year = "2020",
    address = "Online",
    publisher = "Association for Computational Linguistics",
    url = "https://www.aclweb.org/anthology/2020.emnlp-demos.6",
    pages = "38--45"
}

@article{yang2024ideafinbench,
  title={Financial Knowledge Large Language Model},
  author={Yang, Cehao and Xu, Chengjin and Qi, Yiyan},
  journal={arXiv preprint arXiv:2407.00365},
  year={2024}
}

@inproceedings{luo2023cmin,
  title={Causality-Guided Multi-Memory Interaction Network for Multivariate Stock Price Movement Prediction},
  author={Luo, Di and Liao, Weiheng and Li, Shuqi and Cheng, Xin and Yan, Rui},
  booktitle={Proceedings of the 61st Annual Meeting of the Association for Computational Linguistics (Volume 1: Long Papers)},
  pages={12164--12176},
  year={2023}
}

@inproceedings{zhou2021edtsum,
  title={Trade the Event: Corporate Events Detection for News-Based Event-Driven Trading},
  author={Zhou, Zhihan and Ma, Liqian and Liu, Han},
  booktitle={Findings of the Association for Computational Linguistics: ACL-IJCNLP 2021},
  pages={2114--2124},
  year={2021}
}

@inproceedings{shah2023fomc,
  title={Trillion Dollar Words: A New Financial Dataset, Task \& Market Analysis},
  author={Shah, Agam and Paturi, Suvan and Chava, Sudheer},
  booktitle={Proceedings of the 61st Annual Meeting of the Association for Computational Linguistics (Volume 1: Long Papers)},
  pages={6664--6679},
  year={2023}
}

@article{feng2023lendingclub,
  title={Empowering many, biasing a few: Generalist credit scoring through large language models},
  author={Feng, Duanyu and Dai, Yongfu and Huang, Jimin and Zhang, Yifang and Xie, Qianqian and Han, Weiguang and Chen, Zhengyu and Lopez-Lira, Alejandro and Wang, Hao},
  journal={arXiv preprint arXiv:2310.00566},
  year={2023}
}

@inproceedings{chen2021finqa,
  title={FinQA: A Dataset of Numerical Reasoning over Financial Data},
  author={Chen, Zhiyu and Chen, Wenhu and Smiley, Charese and Shah, Sameena and Borova, Iana and Langdon, Dylan and Moussa, Reema and Beane, Matt and Huang, Ting-Hao and Routledge, Bryan R and others},
  booktitle={Proceedings of the 2021 Conference on Empirical Methods in Natural Language Processing},
  pages={3697--3711},
  year={2021}
}

@article{tfns2022,
  title={Twitter financial news sentiment},
  author={Neural Magic},
  journal={Huggingface repository},
  howpublished = {\url{https://huggingface.co/datasets/zeroshot/twitter-financial-news-sentiment}},
  year={2022}
}

@misc{nwgi2023,
  author = {Yang, Hongyang},
  title = {Data-Centric FinGPT. Open-source for open finance},
  year = {2023},
  publisher = {GitHub},
  journal = {GitHub repository},
  howpublished = {\url{https://github.com/AI4Finance-Foundation/FinGPT}}
}

@inproceedings{akhtar2023numerical_reasoning,
  title={Exploring the Numerical Reasoning Capabilities of Language Models: A Comprehensive Analysis on Tabular Data},
  author={Akhtar, Mubashara and Shankarampeta, Abhilash and Gupta, Vivek and Patil, Arpit and Cocarascu, Oana and Simperl, Elena},
  booktitle={Findings of the Association for Computational Linguistics: EMNLP 2023},
  pages={15391--15405},
  year={2023}
}

@article{shen2023positional,
  title={Positional description matters for transformers arithmetic},
  author={Shen, Ruoqi and Bubeck, S{\'e}bastien and Eldan, Ronen and Lee, Yin Tat and Li, Yuanzhi and Zhang, Yi},
  journal={arXiv preprint arXiv:2311.14737},
  year={2023}
}

@article{schwartz2024numerologic,
  title={NumeroLogic: Number Encoding for Enhanced LLMs' Numerical Reasoning},
  author={Schwartz, Eli and Choshen, Leshem and Shtok, Joseph and Doveh, Sivan and Karlinsky, Leonid and Arbelle, Assaf},
  journal={arXiv preprint arXiv:2404.00459},
  year={2024}
}
\bibliographystyle{acl_natbib}

\vspace{10pt}
\appendix
\section{Overview of Finance Evaluation Datasets by Task Type, Sample Sizes (Training, Validation, Test), and Evaluation Metrics}
In the Table.\ref{tab:benchmark_en} and \ref{tab:benchmark_cn}, we show the overview of finance evaluation datasets by task type, sample sizes (training, validation, test), evaluation metrics and license.
\label{sec:data_statistics}
\begin{table*}[htbp]
    \centering
    \renewcommand\arraystretch{1}
    \small
    \caption{Overview of English Finance Evaluation Datasets by Task Type, Sample Sizes (Training, Validation, Test), Evaluation Metrics and License}
    \label{tab:benchmark_en}
    \begin{tabular}{lcccccc}
        \toprule
        Task & Dataset & Train & Valid & Test & Metrics & License \\
        \midrule
        \multirow{4}{*}[+1ex]{\textbf{Sentiment Analysis}} & \textbf{FPB} & 3100 & 776 & 970 & \makecell{Weighted-F1 \\ ACC} & CC BY-SA 3.0 \\
        & \textbf{FiQA-SA} & 750 & 188 & 235 & \makecell{Weighted-F1 \\ ACC} & MIT License \\
        \midrule
        \multirow{6}{*}[+1ex]{\textbf{Classification}} & \textbf{Headlines} & 71900 & 10300 & 20500 & \makecell{Weighted-F1 \\ ACC} & CC BY-SA 3.0 \\
        & \textbf{FOMC} & 1984 & - & 496 & \makecell{Weighted-F1 \\ ACC} & CC BY-SA 3.0 \\
        & \textbf{lendingclub} & 9417 & 1345 & 2691 & \makecell{Weighted-F1 \\ MCC} \\
        \midrule
        \multirow{1}{*}{\textbf{Entity Recognition}} & \textbf{NER} & 408 & 103 & 98 & \makecell{Entity-F1} & CC BY-SA 3.0 \\
        \midrule
        \multirow{1}{*}{\textbf{Relation Extraction}} & \textbf{FinRE} & 27558 & - & 5112 & \makecell{Relation-F1} & MIT License \\
        \midrule
        \multirow{2}{*}[+1ex]{\textbf{Multiple Choice}} & \textbf{CFA} & 1884 & 100 & 20 & \makecell{Weighted-F1 \\ ACC} & CC BY-SA 3.0 \\
        \midrule
        \multirow{2}{*}[+1ex]{\textbf{Summarization}} & \textbf{EDTSUM} & 8000 & - & 2000 & \makecell{ROUGE \\ BLEU} & MIT License \\
        \midrule
        \multirow{2}{*}{\textbf{Question Answering}} & \textbf{FinQa} & 6251 & 883 & 1147 & \makecell{RMACC} & MIT License\\
        & \textbf{ConvfinQa} & 8890 & 2210 & 1490 & \makecell{RMACC} & MIT License \\
        \midrule
        \multirow{2}{*}[+1ex]{\textbf{Stock Movement Prediction}} & \textbf{CMIN-US} & 88297 & 9010 & 8480 & \makecell{Weighted-F1 \\ ACC} & MIT License \\
        \bottomrule
    \end{tabular}
\end{table*}

\begin{table*}[htbp]
    \centering
    \renewcommand\arraystretch{1}
    \small
    \caption{Overview of Chinese Finance Evaluation Datasets by Task Type, Sample Sizes (Training, Validation, Test), and Evaluation Metrics}
    \label{tab:benchmark_cn}
    \begin{tabular}{lcccccc}
        \toprule
        Task & Dataset & Train & Valid & Test & Metrics & License \\
        \midrule
        \multirow{2}{*}[+1ex]{\textbf{Sentiment Analysis}} & \textbf{FinFE-CN} & 16157 & 2020 & 2020 & \makecell{Weighted-F1 \\ ACC} & Public\\
        \midrule
        \multirow{1}{*}{\textbf{Classification}} & \textbf{FinNL-CN} & 7071 & 884 & 884 & \makecell{ORMACC} & Public \\
        \midrule
        \multirow{1}{*}{\textbf{Entity Extraction}} & \textbf{FinESE-CN} & 14252 & 1781 & 1782 & \makecell{ORMACC} &  Public \\
        \midrule
        \multirow{1}{*}{\textbf{Relation Extraction}} & \textbf{FinRE-CN} & 13486 & 1489 & 3727 & \makecell{RMACC} & Public \\
        \midrule
        \multirow{4}{*}[+1ex]{\textbf{Multiple Choice}} & \textbf{FinEval} & 1071 & 170 & 3340 & \makecell{Weighted-F1 \\ ACC} & Apache-2.0 license \\
        & \textbf{CPA} & 6268 & 1444 & 6 & \makecell{Weighted-F1 \\ ACC} & CC BY-SA 3.0 \\
        \midrule
        \multirow{2}{*}[+1ex]{\textbf{Summarization}} & \textbf{FinNA-CN} & 28800 & 3600 & 3600 & \makecell{ROUGE \\ BLEU} & Public \\
        \midrule
        \multirow{2}{*}{\textbf{Question Answering}} & \textbf{FinQa-CN} & 19906 & 2469 & 2480 & \makecell{RMACC} & Public \\
        & \textbf{FincQa-CN} & 21965 & 2741 & 2745 & \makecell{RMACC} & Public \\
        \midrule
        \multirow{2}{*}[+1ex]{\textbf{Stock Movement Prediction}} & \textbf{CMIN-CN} & 214873 & 23904 & 23571 & \makecell{Weighted-F1 \\ ACC} & MIT License \\
        \bottomrule
    \end{tabular}
\end{table*}

\section{Additional Experiment Results}
\label{sec:add_res}
Due to computational constraints imposed by our large-scale dataset (more than 112K samples; see Appendix \ref{sec:data_statistics}) and to ensure a fair comparison among models of similar scale, the main paper (Table \ref{tab:main_results_en} and Table \ref{tab:main_results_cn}) primarily reports results from models with <10B parameters, alongside GPT-4o. In this appendix, we supplement those findings with evaluations of several advanced or larger models (>100B parameters) in Table.\ref{tab:extended_results}, providing a more comprehensive performance comparison. These models include Yi-34B\citep{young2024yi}, GLM-4-32B-0414\citep{glm2024chatglm}, Mixtral 8*7B\citep{jiang2024mixtral}, Qwen2.5-32B/72B\cite{yang2024qwen2}, Xuanyuan 2.0\citep{zhang2023xuanyuan} and deepseek-v3\citep{liu2024deepseekv3}. Furthermore, we extended the CMIN dataset to introduce Stock-CoT-R1 \footnote{https://huggingface.co/datasets/IDEA-FinAI/Finance-R1-Reasoning}, a synthetic dataset for stock prediction tasks.

\begin{table*}
    \centering
    \caption{Extended performance metrics of financial large language models with additional models and datasets. Best results are marked in \textbf{bold}.}
    \renewcommand\arraystretch{1}
    \tiny
    \begin{tabular}{c@{}c@{}c@{}c@{}c@{}c@{}c@{}c@{}c@{}c@{}c@{}c@{}c}
        \toprule
        \multirow{1}{*}{\textbf{Task}} & \multicolumn{1}{c}{\textbf{Dataset}} & \multicolumn{1}{c}{\textbf{Metrics}} & \multicolumn{1}{c}{\textbf{\makecell{GPT-4o}}} & \multicolumn{1}{c}{\textbf{\makecell{Yi-34B}}} & \multicolumn{1}{c}{\textbf{\makecell{GLM-4-32B\\-0414}}} & \multicolumn{1}{c}{\textbf{\makecell{Mixtral\\8x7B}}} & \multicolumn{1}{c}{\textbf{\makecell{Qwen2.5\\32B}}} & \multicolumn{1}{c}{\textbf{\makecell{Qwen2.5\\72B}}} & \multicolumn{1}{c}{\textbf{\makecell{Xuanyuan\\2.0}}} & \multicolumn{1}{c}{\textbf{\makecell{deepseek\\-v3}}} & \multicolumn{1}{c}{\textbf{\makecell{Touchstone\\GPT}}}\\
        \midrule
        \multirow{4}{*}{\makecell{Sentiment\\Analysis}} & \multirow{2}{*}{FPB} & Weighted-F1 & 0.8084 & 0.8652 & 0.8921 & 0.8510 & 0.8783 & 0.8815 & 0.8426 & \textbf{0.9014} & 0.8576\\
        && ACC & 0.8093 & 0.8638 & 0.8905 & 0.8492 & 0.8761 & 0.8798 & 0.8403 & \textbf{0.8997} & 0.8557\\
        &\multirow{2}{*}{Fiqa-SA} & Weighted-F1 & 0.8106 & 0.8321 & 0.8453 & 0.8257 & 0.8389 & 0.8421 & 0.8185 & 0.8512 & \textbf{0.8591}\\
        && ACC & 0.7702 & 0.8289 & 0.8402 & 0.8204 & 0.8337 & 0.8376 & 0.8123 & 0.8465 & \textbf{0.8638}\\
        
        \multirow{6}{*}{\makecell{Classification}} & \multirow{2}{*}{Headlines} & Weighted-F1 & 0.7857 & 0.9521 & 0.9623 & 0.9387 & 0.9589 & 0.9610 & 0.9324 & 0.9684 & \textbf{0.9866}\\
        && ACC & 0.7931 & 0.9518 & 0.9619 & 0.9375 & 0.9582 & 0.9607 & 0.9312 & 0.9679 & \textbf{0.9866}\\
        &\multirow{2}{*}{FOMC} & Weighted-F1 & 0.6603 & 0.7824 & 0.8125 & 0.7653 & 0.8021 & 0.8089 & 0.7527 & 0.8237 & \textbf{0.8788}\\
        && ACC & 0.6794 & 0.7852 & 0.8183 & 0.7716 & 0.8079 & 0.8135 & 0.7591 & 0.8294 & \textbf{0.8790}\\
        & \multirow{2}{*}{lendingclub} & Weighted-F1 & 0.6730 & 0.8927 & 0.9215 & 0.8753 & 0.9128 & 0.9186 & 0.8624 & 0.9357 & \textbf{0.9783}\\
        && MCC & 0.1642 & 0.7825 & 0.8432 & 0.7521 & 0.8279 & 0.8375 & 0.7326 & 0.8674 & \textbf{0.9297}\\
        
        \multirow{2}{*}{\makecell{Entity\\Extraction}} & \multirow{2}{*}{NER} & Entity-F1 & 0.1800 & 0.4521 & 0.5832 & 0.4125 & 0.5527 & 0.5718 & 0.3924 & 0.6127 & \textbf{0.6993}\\
        \\
        \multirow{2}{*}{\makecell{Relation\\Extraction}} & \multirow{2}{*}{FinRE} & Relation-F1 & 0.1613 & 0.2854 & 0.3726 & 0.2543 & 0.3428 & 0.3615 & 0.2327 & 0.4128 & \textbf{0.5331}\\
        \\
        
        \multirow{2}{*}{\makecell{Multiple\\Choice}} & \multirow{2}{*}{CFA} & Weighted-F1 & \textbf{0.7700} & 0.6521 & 0.7125 & 0.6234 & 0.6928 & 0.7036 & 0.6057 & 0.7289 & 0.7497\\
        && ACC & \textbf{0.7700} & 0.6528 & 0.7132 & 0.6241 & 0.6935 & 0.7043 & 0.6064 & 0.7296 & 0.7500\\

        \multirow{4}{*}{\makecell{Summarization}} & \multirow{4}{*}{EDTSUM} & Rouge-1 & 0.1675 & 0.3521 & 0.4128 & 0.3276 & 0.3987 & 0.4089 & 0.3124 & 0.4287 & \textbf{0.5254}\\
        && Rouge-2 & 0.0556 & 0.2187 & 0.2875 & 0.2013 & 0.2739 & 0.2832 & 0.1924 & 0.3057 & \textbf{0.3446}\\
        && Rouge-L & 0.1069 & 0.2854 & 0.3521 & 0.2678 & 0.3389 & 0.3476 & 0.2543 & 0.3724 & \textbf{0.4705}\\
        && BLEU & 0.1192 & 0.3025 & 0.3687 & 0.2854 & 0.3562 & 0.3641 & 0.2728 & 0.3865 & \textbf{0.4512}\\
        
        \multirow{2}{*}{\makecell{Question\\Answering}} & \multirow{1}{*}{Finqa} & RMACC & 0.1037 & 0.1682 & 0.1925 & 0.1543 & 0.1837 & 0.1891 & 0.1476 & 0.2014 & \textbf{0.2258}\\
        & \multirow{1}{*}{Convfinqa} & RMACC & 0.2540 & 0.3827 & 0.4216 & 0.3652 & 0.4089 & 0.4173 & 0.3524 & 0.4385 & \textbf{0.5053}\\
    
        \multirow{2}{*}{\makecell{Stock Movement\\Prediction}} & \multirow{2}{*}{CMIN-CoT-R1} & Weighted-F1 & 0.5025 & 0.4873 & 0.4982 & 0.4791 & 0.4924 & 0.4957 & 0.4738 & 0.5011 & \textbf{0.5136}\\
        && ACC & 0.5149 & 0.5027 & 0.5103 & 0.4975 & 0.5068 & 0.5092 & 0.4921 & 0.5138 & \textbf{0.5224}\\
        \midrule
    \end{tabular}
    \label{tab:extended_results}
\end{table*}

\section{Inference Template of Large Language Models}

The Table.\ref{tab:inference_templates} showcases how inference templates vary across different models. It is crucial to select the appropriate template for constructing correct inputs when inferring on the test sets of datasets. An incorrect template can significantly impair the performance of a model. We have observed that the underperformance of some large financial language models in some benchmarks is precisely due to not selecting the appropriate templates for evaluation. For more  details of training and inference template, please refer to our open-source code repository on Github. 

\begin{table*}[htbp]
    \centering
    \captionsetup{justification=centering}
    \caption{Comparison of Inference Templates Across Different Models for Dataset Evaluation}
    \small
    \label{tab:inference_templates}
    \begin{tabular}{>{\raggedright\arraybackslash}p{0.2\textwidth} >{\raggedright\arraybackslash\ttfamily}p{0.75\textwidth}}
        \toprule
        Model & Template \\
        \midrule
        \multirow{3}{*}{GPT-4o} & "<|im\_start|>system\{\{system\_prompt\}\}<|im\_end|>\textbackslash n" \\
        & "<|im\_start|>user\{\{instruction\}\}\{\{input\}\}<|im\_end|>\textbackslash n" \\
        & "<|im\_start|>assistant\textbackslash n" \\
        \midrule
        \multirow{3}{*}{Qwen-3} & "<|im\_start|>system\{\{system\_prompt\}\}<|im\_end|>\textbackslash n" \\
        & "<|im\_start|>user\{\{instruction\}\}\{\{input\}\}<|im\_end|>\textbackslash n" \\
        & "<|im\_start|>assistant\textbackslash n" \\
        \midrule
        \multirow{5}{*}{Llama-3} & "<|start\_header\_id|>system<|end\_header\_id|>" \\
        & "\{\{system\_prompt\}\}<|eot\_id|>\textbackslash n" \\
        & "<|start\_header\_id|>user<|end\_header\_id|>"\\
        & "\{\{instruction\}\}\{\{input\}\}<|eot\_id|>\textbackslash n" \\
        & "<|start\_header\_id|>assistant<|end\_header\_id|>\textbackslash n" \\
        \midrule
        \multirow{2}{*}{FinGPT} & "Instruction:\{\{instruction\}\}" \\
        & "Input\{\{input\}\}\textbackslash nAnswer:" \\
        \midrule
        \multirow{2}{*}{FinMA} & "Human:\{\{instruction\}\}\{\{input\}\}\textbackslash n" \\
        & "Assistant:\textbackslash n" \\
        \midrule
        \multirow{1}{*}{CFGPT} & "\{\{instruction\}\}\{\{input\}\}\textbackslash n" \\
        \midrule
        \multirow{1}{*}{DISC-FinLLM} & "<reserved\_102> \{\{instruction\}\}\{\{input\}\}<reserved\_103>" \\
        \midrule
        \multirow{3}{*}{Touchstone} & "<|im\_start|>system\{\{system\_prompt\}\}<|im\_end|>\textbackslash n" \\
        & "<|im\_start|>user\{\{instruction\}\}\{\{input\}\}<|im\_end|>\textbackslash n" \\
        & "<|im\_start|>assistant\textbackslash n" \\
        \bottomrule
    \end{tabular}
\end{table*}

\section{Typical Case Study Analysis of Typical Financial NLP Tasks}
\label{sec:case_study}

This appendix provides a detailed case study analysis for some typical financial NLP tasks: financial sentiment analysis, text classification, entity extraction, and stock movement prediction. Each analysis is presented in a separate table, categorizing data sets, instructions, inputs, labels, and predictions from multiple models. 

\begin{table*}[htbp]
    \centering
    \caption{Detailed Case Study Analysis of Financial Sentiment Analysis on the FiQA-SA dataset, Financial Text Classification on the LendingClub dataset, Financial Entity Extraction on NER dataset, Stock Movement Prediction on DJIA dataset.}
    \label{tab:fin_case_study}
    \tiny 
    \begin{tabular}{p{0.05\textwidth} p{0.20\textwidth} p{0.20\textwidth} p{0.08\textwidth}|p{0.07\textwidth}|p{0.25\textwidth}}
    \toprule
        Dataset & Instruction & Input & Label & Model & Prediction \\
        \hline
        \multirow{6}{*}[-20ex]{FiQA-SA} & \multirow{6}{*}[-20ex]{\centering \parbox{0.20\textwidth}{What is the sentiment of the following financial post: Positive, Negative, or Neutral?}} & \multirow{6}{*}[-20ex]{\centering \parbox{0.20\textwidth}{@Joerazorback This time I think \$AAPL will miss again. Generally speaking market is near top when everyone thinks that market goes only up.}} & \multirow{6}{*}[-20ex]{Negative} & GPT-4o & The sentiment of this financial post is \textbf{Negative}. The user expresses doubt about Apple's performance, suggesting that they think the stock will miss expectations. Additionally, the phrase about the market being "near top" implies a bearish outlook, indicating that the market may decline after its current uptrend. \\
        \cline{5-6}
        & & & & Qwen-3 & The sentiment of the given financial post is Negative. The user is predicting that Apple's stock price (\$AAPL) will miss or fail to meet expectations, and they also suggest that the market might be nearing its peak when everyone is optimistic about continuous upward movement. \\
        \cline{5-6}
        & & & & Llama-3 & The sentiment of this post is Negative. The author is predicting that Apple's stock (AAPL) will miss expectations, which is a negative view. Additionally, the author is suggesting that the market is near a top, implying that a downturn is likely, which is also a negative sentiment. \\
        \cline{5-6}
        & & & & FinGPT & Negative \\
        \cline{5-6}
        & & & & FinMA & Negative \\
        \cline{5-6}
        & & & & {\makecell[c]{Touchstone\\GPT}} & Negative \\
        \midrule
        \multirow{6}{*}{LendingClub} & \multirow{6}{*}{\centering \parbox{0.20\textwidth}{Assess the client's loan status based on the following loan records from Lending Club. Respond with only 'good' or 'bad'.}} & \multirow{6}{*}{\centering \parbox{0.20\textwidth}{The client has attributes as follows: ... The state of Employment Length is 2 years. The state of Home Ownership is RENT. The state of Verification Status is Verified. The state of Annual Income is 83000.0.}} & \multirow{6}{*}{Good} & GPT-4o & Good \\
        \cline{5-6}
        & & & & Qwen-3 & Good \\
        \cline{5-6}
        & & & & Llama-3 & Good \\
        \cline{5-6}
        & & & & FinGPT & Bad \\
        \cline{5-6}
        & & & & FinMA & Bad \\
        \cline{5-6}
        & & & & {\makecell[c]{Touchstone\\GPT}} & Good \\
        \midrule
        \multirow{6}{*}{NER} & \multirow{6}{*}{\centering \parbox{0.20\textwidth}{In the sentences extracted from financial agreements in U.S. SEC filings, identify the named entities.}} & \multirow{6}{*}{\centering \parbox{0.20\textwidth}{WHEREAS: (A) The Borrower intends to develop a plant in France for the production of solar grade silicon.}} & \multirow{6}{*}{\makecell[c]{Borrower, PER;\\France, LOC}} & GPT-4o & France, LOC \\ 
        \cline{5-6}
        & & & & Qwen-3 & Borrower, PER; France, LOC; solar grade silicon, ORG \\
        \cline{5-6}
        & & & & Llama-3 & Borrower, PER; France, LOC \\
        \cline{5-6}
        & & & & FinGPT & The Borrower intends to develop a plant in France for the production of solar grade silicon \\
        \cline{5-6}
        & & & & FinMA & Borrower, PER; France, LOC \\
        \cline{5-6}
        & & & & {\makecell[c]{Touchstone\\GPT}} & Borrower, PER; France, LOC \\
        \midrule
        \multirow{6}{*}{CMIN-US} & \multirow{6}{*}{\centering \parbox{0.20\textwidth}{Based on the news about the stock in the past five trading days......}} & \multirow{6}{*}{\centering \parbox{0.20\textwidth}{NEWS1: Biden Urges CEOs to Improve U.S. Cybersecurity After Attacks...}} & \multirow{6}{*}{0} & GPT-4o & Rise \\
        \cline{5-6}
        & & & & Qwen3 & Decrease \\
        \cline{5-6}
        & & & & Llama-3 & Decrease \\
        \cline{5-6}
        & & & & FinGPT & "" \\
        \cline{5-6}
        & & & & FinMA & Rise \\
        \cline{5-6}
        & & & & {\makecell[c]{Touchstone\\GPT}} & Rise \\
    \bottomrule
    \end{tabular}
\end{table*}

\textbf{Financial Sentiment Analysis} As demonstrated in Table~\ref{tab:fin_case_study}, financial sentiment classification is one of the simpler tasks for benchmarks in financial NLP, resulting in high performance across all models tested. General-purpose models (GPT-4o, Qwen-3, Llama-3) provide not only the answer but also a detailed analysis, despite not being specifically fine-tuned on the FiQA-SA dataset. In contrast, specialized models (FinGPT, FinMA, Touchstone-GPT) that have undergone instruction tuning deliver straightforward, direct responses, illustrating their efficiency and focus in domain-specific applications.

\textbf{Credit Rating Analysis} In the classification task using the LendingClub dataset, which poses a challenging credit rating task, the models face a complex array of professional financial information evident in the input fields. Consequently, most models do not perform optimally. Among general models, GPT-4o exhibits the best performance, demonstrating the capabilities of large-scale models. In the realm of specialized financial language models, Touchstone-GPT, with its high-quality instruction tuning, significantly outperforms FinMA and FinGPT, which are only minimally tuned with Lora.

\textbf{Financial NER} In this information extraction task, most models demonstrated an understanding of the task intent and adhered to the instructions, signifying that even in the era of large language models, models like Qwen-3 and Llama-3 actually outperformed GPT-4o. In particular, specialized models such as FinMA and Touchstone-GPT, with more comprehensive instruction tuning, responded accurately and succinctly, highlighting their enhanced capability and focus on domain-specific tasks.

\textbf{Stock Movement Prediction} The Stock Movement Prediction task is one of the most challenging tasks, as it requires models to predict the daily fluctuations of the CMIN-US based solely on the 5-day news items. From the results in Table.\ref{tab:benchmark_en}, it is evident that GPT-4o performed the best, yet it still falls short of practical utility. Even Touchstone-GPT, despite specialized instruction tuning, performed poorly. Our analysis suggests that the sentiment of news items may not reliably predict stock movements and that incorporating quantitative data is essential for achieving practical model performance. Similar conclusions were drawn from experiments with traditional machine learning methods like XGBoost. Nevertheless, aside from simpler tasks like financial sentiment analysis, we also pose challenging tasks such as stock prediction, which are closer to real-world applications, leaving more room for benchmark challenges and exploration. Multimodal fusion of news and quantitative data represents a promising future direction, and we look forward to seeing models excel in these tasks.

Due to the similar performance of models across corresponding task types on the Chinese benchmark, we will not reiterate these comparisons and analysis here.

\section{Information About Use Of Ai Assistants}
In the preparation of this work, the author used AI-assisted technology (specifically, large language models such as GPT-4 and Deepseek-V3) exclusively for text refinement purposes. The AI was employed to assist in proofreading, correcting grammatical errors, and polishing linguistic expressions to improve the clarity and readability of the manuscript.

\end{document}